%% file: main.tex
\documentclass[acmtog]{acmart}
\acmSubmissionID{1234}
\renewcommand\footnotetextcopyrightpermission[1]{}

\usepackage{booktabs} 

\usepackage{enumitem} 
\usepackage{xcolor} 
\usepackage{colortbl} 

\usepackage[ruled,vlined,noend]{algorithm2e} 

\SetCommentSty{mycommfont}

\usepackage{appendix}
\input{src/code}

\SetAlFnt{\small}
\SetAlCapFnt{\small}
\SetAlCapNameFnt{\small}
\SetAlCapHSkip{0pt}

\newcommand{\ShowOrHide}[1]{} 
\definecolor{darkgreen}{rgb}{0.0, 0.25, 0.0}

\newcommand{\OurPool}{3Pool}
\newcommand{\Assistant}{CueTip}

\newcommand{\Simi}{\ensuremath{f_e}}
\newcommand{\Sim}{\ensuremath{f}}
\newcommand{\Q}{\ensuremath{Q}}
\newcommand{\E}{\ensuremath{E}}
\newcommand{\X}{\ensuremath{\mathbf x}}
\newcommand{\T}{\ensuremath{\theta}}
\newcommand{\A}{\ensuremath{\pi}}
\newcommand{\AL}{\ensuremath{\Pi}}

\newcommand{\AN}{\ensuremath{\tilde{\pi}}}
\newcommand{\Ev}{\ensuremath{L}}
\newcommand{\Evi}{\ensuremath{\Ev_i}}
\newcommand{\cp}[4]{\ensuremath{p(#1 \left \vert \vphantom{#1#2} 
\right. #2, #3, #4)}}
\newcommand{\cpp}[3]{\ensuremath{p(#1 \left \vert \vphantom{#2} 
\right. #2, #3)}}
\newcommand{\vn}{\ensuremath{\tilde{\mathbf p}}}
\newcommand{\LM}{\ensuremath{\ell}}

\newcommand{\Apm}{\ensuremath{\pi_\text{pm}}}
\newcommand{\ALpm}{\ensuremath{\Pi_{\text{pm}}}}
\newcommand{\ANpm}{\ensuremath{\tilde{\A}_\text{pm}}}

\newcommand{\Ag}{\ensuremath{\pi_\text{g}}}
\newcommand{\ALg}{\ensuremath{\Pi_\text{g}}}
\newcommand{\ANg}{\ensuremath{\tilde{\A}_\text{g}}}

\newcommand{\pVar}[2]{\colorbox{#1}{#2}}
\newcommand{\pgvar}[1]{\pVar{green}{\{#1\}}}
\newcommand{\pfVar}[2]{\colorbox{#1}{\framebox{#2}}}

\begin{document}

\title{CueTip: An Interactive and Explainable Physics-aware Pool Assistant}

\author{Sean Memery}
\email{s.memery@ed.ac.uk}
\orcid{0009-0004-2437-5154}
\affiliation{%
  \institution{University of Edinburgh}
  \city{Edinburgh}
  \country{United Kingdom}
}

\author{Kevin Denamganai}
\email{kevin.denamganai@ed.ac.uk}
\affiliation{%
  \institution{University of Edinburgh}
  \city{Edinburgh}
  \country{United Kingdom}
}

\author{Jiaxin Zhang}
\email{jiaxijzhang@global.tencent.com}
\affiliation{%
  \institution{Lightspeed Studios}
  \city{London}
  \country{United Kingdom}
}

\author{Zehai Tu}
\email{zehaitu@global.tencent.com}
\affiliation{%
  \institution{Lightspeed Studios}
  \city{London}
  \country{United Kingdom}
}

\author{Yiwen Guo}
\email{guoyiwen89@gmail.com}
\affiliation{%
  \institution{Independent Researcher}
  \city{Beijing}
  \country{China}
}

\author{Kartic Subr}
\email{k.subr@ed.ac.uk}
\affiliation{%
  \institution{University of Edinburgh}
  \city{Edinburgh}
  \country{United Kingdom}
}

\begin{teaserfigure}
	\begin{center}		
        \includegraphics[width=0.95\columnwidth]{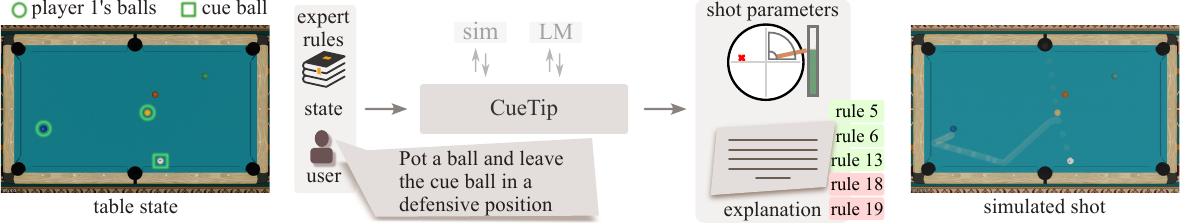} \\         
	\caption{\Assistant\ provides interactive pool coaching through physically-grounded, explainable decision-making. Given a query requesting a defensive shot with ball potting, \Assistant\ identifies a solution that pots the yellow ball while positioning the cue ball adjacent to the blue ball, as visualized in the imulated shot image, restricting the opponent's freedom. 
	\Assistant's explanation for why the user should take this shot is rooted in expert-provided rules, in this case: rule $5$ (defensive positioning) contributes positively, while rule $19$ (spin complexity) indicates increased difficulty (cf. Appendix D for more details).}
    \label{fig:teaser}
	\end{center}
\end{teaserfigure}

\begin{abstract}
We present an interactive and explainable automated coaching assistant called \Assistant\ for a variant of pool/billiards. \Assistant's novelty lies in its combination of three features: 
a natural-language interface, an ability to perform contextual, physics-aware reasoning, and that its explanations are rooted in a set of predetermined guidelines developed by domain experts.  We instrument a physics simulator so that it generates event traces in natural language alongside traditional state traces. Event traces lend themselves to interpretation by language models, which serve as the interface to our assistant. We design and train a neural adaptor that decouples tactical choices made by \Assistant\ from its interactivity and explainability allowing it to be reconfigured to mimic any pool playing agent. Our experiments show that \Assistant\ enables contextual query-based assistance and explanations while maintaining the strength of the agent in terms of win rate (improving it in some situations). The explanations generated by \Assistant\ are physically-aware and grounded in the expert rules and are therefore more reliable.
\end{abstract}

%
%

\ccsdesc[500]{Computing methodologies~Artificial intelligence}
\ccsdesc[500]{Computing methodologies~Knowledge representation and reasoning}
\ccsdesc[300]{Computing methodologies~Natural language processing}
\ccsdesc[300]{Computing methodologies~Physical simulation}
\ccsdesc[300]{Computing methodologies~Intelligent agents}

%
%

\keywords{Physical Reasoning, Planning, Explainability, Language Models}

\maketitle

\input{src/introduction}

\input{src/background}

\input{src/method}
\input{src/experiments}
\input{src/table}
\input{src/results}

\input{src/conclusion}
\input{src/acknowledgments}

\input{src/image_pages}

\bibliographystyle{ACM-Reference-Format}
\bibliography{main}

\input{src/appendix}

\end{document}

%% file: src/code.tex
\usepackage{geometry}
\usepackage{graphicx}

\usepackage{amsmath}
\usepackage{listings}
\usepackage{xcolor}
\usepackage{enumitem}

\definecolor{codegreen}{rgb}{0,0.6,0}
\definecolor{codegray}{rgb}{0.5,0.5,0.5}

\definecolor{backcolour}{RGB}{245,248,250}
\definecolor{emph}{RGB}{166,88,53}
\definecolor{nightblue}{RGB}{9,49,105}
\definecolor{keywords}{RGB}{207,33,46}
\definecolor{lightpurple}{RGB}{130,81,223}

\lstdefinestyle{mystyle}{
    backgroundcolor=\color{backcolour},   
    commentstyle=\color{codegreen},
    keywordstyle=\color{keywords},
    stringstyle=\color{nightblue},
    basicstyle=\fontsize{7}{8}\ttfamily,
    breakatwhitespace=true,         
    breaklines=true,                 
    captionpos=b,                    
    keepspaces=true,                 
    numberstyle=\tiny\color{codegray},
    numbersep=2pt,                  
    showspaces=false,                
    showstringspaces=false,
    showtabs=false,                  
    tabsize=2,
    emph={dspy},
    emphstyle={\color{lightpurple}},
    linewidth=1\columnwidth,
    frame=tb,    
    xrightmargin=0pt,
    xleftmargin=0.23cm,
    numbers=left,
    aboveskip=0.2cm,
    belowskip=0.1cm,
}

%% file: src/introduction.tex
\section{Introduction}
\label{sec:introduction}

Imagine interacting and reasoning in natural language with an automatic assistant about shot options for a game of pool. Now, given a description of balls on a table, imagine querying a language model (LM) to predict their positions after a specific shot is taken. While LMs are excellent at conversational exchange, they are inherently ill-suited to reason about dynamics in a physical system. Even if the initial state, shot dynamics, and all physical parameters can be described exactly, the LM is likely to hallucinate predictions that are physically implausible.  

We use an \emph{off-the-shelf} LM and a physics simulator~\citep{Kiefl2024} in closed loop to address the above problem. Rather than attempting to imbue gargantuan LMs with knowledge of pool-table physics, our insight is to instrument a physical simulator for pool to output a natural language trace composed of a few simple events.  We translate this insight into \Assistant\ --- a natural language assistant for a variant of pool. We demonstrate \Assistant's ability to provide contextual help based on a user query specific to a given state of the table, by suggesting relevant shots along with explanations. An overview of this process is provided in Figure~\ref{fig:teaser}. The assistant is also adapted to empirically mimic the shot-selection strategy of any pool playing agent. \Assistant\ accepts a set of predefined expert rules, i.e. \textit{heuristics}, that are used to evaluate the quality of a shot for a given state. The explanations generated by the assistant are rooted in these rules. 

There is considerable interest in reasoning about physics with LMs using strategies such as training or fine-tuning ~\citep{Liu2022MindsEG,wang2023newtonlargelanguagemodels}, using LMs to perform planning ~\citep{valmeekam2023planning,feng2024layoutgpt,huang2023inner}, or pure prompt engineering \citep{Polverini_2024, Yeadon_2024}. Our approach could be categorized as using LMs to help planning \citep{kambhampati2024LLM-Modulo}. That is, we devise a mechanism where the LM recommends plans in consultation with a simulator but the tuning of that plan is domain-specific. Finally, the tuned plan is explained by the LM with respect to relevant domain-specific guidelines. Our design of \Assistant\ is adaptable and supports a wide range of queries, it can be tuned easily to mimic an agent (used for determining shots) and accommodate additional expert rules (used as guidelines for shot suggestion and explanations).   

We introduce an interactive and explainable natural-language assistant for pool. Within this context, our contributions are: 
\begin{enumerate} 
    [leftmargin=2em]
    \item \emph{Physics awareness}: we build contextual, physics-aware reasoning into the assistant;
    \item \emph{Explainability}: we develop a mechanism for generating reliable explanations, grounded in   predetermined domain-expert rules;
    \item \emph{Modularity}: we decouple the assistant from the underlying agent at the cost of pretraining a small multilayer perceptron network-- which serves as a neural surrogate of the agent; and
    \item \emph{Uncertainty awareness}: our mechanism accommodates for uncertainty around the value of shots.
\end{enumerate}

\begin{figure*}[t]
    \centering
    \includegraphics[width=0.99\textwidth]{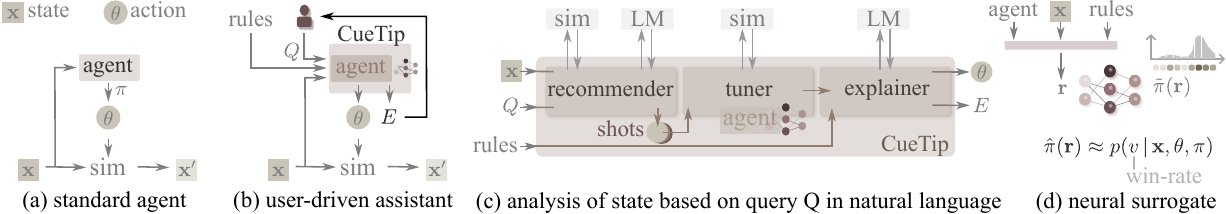}
    \caption{\label{fig:overview} A standard pool agent (a) makes a decision on which shot \T\ to play for a given state \X\ of the table. We propose an interactive assistant (b) that contains an embodiment of an agent, but in addition is able to tune the choice of shot to user input $Q$. Our assistant produces \T\ and an explanation of this choice \T\ that is grounded in some set of rules defined by domain experts. An overview of \Assistant\ and its interactions with simulator and language model are shown in (c). Instead of using the agent directly within \Assistant, we train a multilayer perceptron network surrogate (d) to mimic any given agent statistically.}
\end{figure*}

%% file: src/background.tex
\section{Background}
\label{sec:background}

\subsection{Related Work}
\label{sec:related-work}

\paragraph{\textbf{Pool AI}}
Pool has a long history of use as a playground for artificial intelligence research. There was a prominent trend on developing performant agents for pool \citep{Sang1994AutomatingSU, Alian2003RobosharkA, Smith2007PickPocketAC, Greenspan2008TowardAC}. Many of these agents competed in the Computational Pool Tournament \citep{Archibald2010ComputationalPA} including CUECARD \citep{Archibald2009AnalysisOA}. Few of these works emphasise explainability, although there is a recent impetus in this direction \citep{Fragkiadaki2015LearningVP, Tung2019TowardHB}. There is some interest in mimicking human decision-making through feature analysis of expert games and a learned network, resulting in improved long-term planning~\citep{Tung2019TowardHB}. We similarily develop a scheme to build an empirical surrogate of a pool-playing agent, but additionally imbue it with interactivity and explainability.

\paragraph{\textbf{Explainable AI and LMs}}

The topic of explainability in relation to LMs has become increasingly popular \citep{zhao2023explainabilitylargelanguagemodels, Luo2024FromUT}. 
Explainability has assumed an important role given an increased appreciation of the limitations of LMs, for example in planning and reasoning~\citep{kambhampati2024LLM-Modulo} or amplifying biases~\citep{gallegos2024BiasAndFairness}.
Their inability to provide explanations~\citep{turpin2024unfaithfulExplanationsInCoT,lanham2023measuringFaithfulness} has led to innovative schemes such as local explanations \citep{Ferrando2022MeasuringTM, Chuang2024LargeLM} (where an LM's output is explained, for example,  through feature attribution), perturbation methods \citep{Lundberg2017AUA} (features of the input are removed to observe changes in output), and gradient-based methods \citep{Sundararajan2017AxiomaticAF} (gradient accumulation is utilised to measure the importance of specific inputs). 
These methods measure relations between features but have low generalization across domains and are not well suited to online decision-making. 

\paragraph{\textbf{LMs and Physics Environments}}

Physical reasoning is an inherently out-of-domain problem for LMs, however many works have attempted to leverage their next-word prediction capabilities within physical environments. 
Following the emergence of math-word problems to benchmark LMs on their math reasoning capabilities~\citep{cobbe2021GSM8K,hendrycks2021MATH,lu2024mathvista}, ~\citep{ding2023PhysQA} proposed the \textit{PhysQA} dataset consisting of physics-word problems to evaluate their physics reasoning capabilities. This form of evaluation is limited, though, as it does not account for dynamic physics interactions.
This is a gap that Mind's Eye \citep{Liu2022MindsEG} starts to address by incorporating the MuJoCo physics simulator~\citep{todorov2012mujoco} to inform question answering via fine tuning of a LM but it does not utilise the simulator to ground the LM's reasoning. 
\citep{Lyu2024AdaptingWL} improves upon the latter by using a tool integration of the MuJoCo physics simulator for LMs being tested on various scientific problems.
Some examples that blend online reasoning and dynamic physics utilise a Minecraft environment \citep{Wang2023VoyagerAO, fan2022minedojobuildingopenendedembodied}, but they typically evaluate LM based agents performance on more general tasks such as exploration and skill acquisition, as opposed to specifically focusing on physics reasoning and understanding. 
Following investigations of intuitive physics~\citep{kubricht2017intuitive} in human beings~\citep{baillargeon1985object,baillargeon1995physical,spelke1992origins,kim1999perception} and machine learning~\citep{battaglia2013simulation,battaglia2016interaction,watters2017visual,smith2019modeling,piloto2022intuitive}, \citep{jassim2024graspnovelbenchmarkevaluating} proposes the \textit{GRASP} benchmark to evaluate more systematically intuitive physics understanding in multimodal LMs.
Results stress a lack of perceptual understanding as a major shortcoming, that ~\citep{balazadeh2024syntheticvisiontrainingvisionlanguage} proposes to improve upon via simulation-to-reality fine-tuning of a modular, specialised visual LM. 
These studies take place in open-loop contexts, i.e. without execution of actions by the LM in the environment. 
Recently, building upon ~\citep{voudouris2022evaluating}, which solely focused on improving the ecological validity of evaluating \textit{object permanence}, \citep{Mecattaf2024ALL} proposed the LLM-AAI benchmark to measure multiple but limited intuitive physics items in several LMs with execution of actions in ecologically valid environments, which are shown to have internal validity towards precisely evaluating intuitive physics understanding and reasoning.
It remains to propose similar environments for all intuitive physics items.
Some works have incorporated the physics simulator in closed-loop reasoning along with an LM \citep{memery2024simlmlanguagemodelsinfer, cherian2024llmphycomplexphysicalreasoning}. These results highlight their unsuitability due to a lack of physical `common-sense' coupled with an ineptness in interpreting numerical simulation traces.

\paragraph{\textbf{Learning from Domain Expert Rules}}

Domain expert rules played a foundational role in artificial intelligence through Expert Systems \citep{LINDSAY1993209, Ryan2017-rx, Yanase2019ASS}, which encoded domain knowledge as explicit if-then statements to emulate expert decision-making. While traditional expert systems faced scalability challenges, modern approaches are exploring the integration of expert knowledge with LMs to combine structured reasoning with natural language capabilities \citep{Arsevska2016MonitoringDO, Louie2024RoleplaydohED, SilvaRodrguez2023AFL}.
This integration offers potential advantages: domain heuristics can provide reliable constraints on LM outputs, while LMs can make expert knowledge more flexible and generalizable. These approaches suggest promising directions for developing systems that balance structured domain expertise with natural language understanding.

\subsection{Setup}
\label{sec:setup}
Let  $\X\in\mathcal X$ represent the state of a pool table  and $\T \in \Theta$ represent shot parameters $(v, \alpha, \beta, a, b)$ consisting of the velocity of the cue stick, $v \in [0, 5]$, its angles (azimuth, $\alpha \in [ 0,360 ]$, and elevation, $\beta \in [ 0,90 ]$) and the point of contact on the cue ball (for imparting spin), $(a,b) \in [-0.5, 0.5]^2$. A standard pool agent $\A: \mathcal X \rightarrow \Theta$ maps a given state of the table to a shot (ideally an optimal play) for that state. We assume that we are given a simulation function  $\Sim: \mathcal X \times \Theta \rightarrow \mathcal X$, so that a shot by a standard agent takes a table state \X\ to  $\Sim(\X, \A(\X))$.

\paragraph{\textbf{\OurPool: A Simplified pool game}}
\label{sec:pool-rules}

We define a variant of 9-ball pool for two players, \OurPool, where the table contains a white (cue) ball and 6 colour balls. Each player is assigned a pre-determined set of 3 colour balls. The game begins with a state \X\ where the balls are in random positions. A player wins when they have pocketed their assigned set of three balls. At each turn, a player (say Alice) strikes the cue ball with a stick according to parameters \T\ taking the table's state to $\Sim(\X,\T)$. If the shot results in one of Alice's balls being pocketed without committing a foul, then Alice may play again; otherwise Alice's opponent plays the next shot. Alice's shot is deemed a foul if: the cue ball makes first contact with her opponent's ball; if the cue ball makes contact with no ball; or if the cue ball is pocketed. Simulating a \emph{complete game} involves executing a sequence of shots by Alice and her opponent until one of them wins (pockets their balls). We adapt an existing simulator~\citep{Kiefl2024} to build a physics-based environment which we call \OurPool.

\paragraph{\textbf{Rules to evaluate shots}}
\label{sec:pool-heuristics}
We adopt a set of rules from the textbook \citep{alciatore2004illustrated}, where domain experts detail heuristics to evaluate the state of a table resulting from a shot. Some rules are assessments of difficulties associated with certain configurations within a state. We choose $m=29$ rules (listed in Appendix A within supplemental material) that apply to \OurPool, representing them as $\mathcal R = \{ R_i, r_i \}, i=1,2,3,\cdots, m$ where  $R_i\in\mathcal L$ is a description of the rule in natural language and $r_i: \mathcal X \times \Theta \rightarrow \mathbb [0,1]$  is a rule-evaluation function that evaluates the relevance of the rule to a state-shot pair. 

\paragraph{\textbf{Agents evaluated}}
\label{sec:poolmaster}
Poolmaster \citep{Dussault2006OptimizationOA}, denoted as $\Apm$, is a tournament-winning pool agent that optimizes shot selection through position-based value estimation. The agent uses an objective function that considers target ball potting, cue ball positioning, and shot legality. Shot evaluation considers both direct and indirect shots through hand-crafted difficulty coefficients, with indirect shots transformed via table mirroring for unified assessment.
Shot selection is performed through a grid-based search, where the table is discretized into a $G \in \{15, 30\}$ grid. For each position, the agent evaluates discretized shot parameters to identify positions that maximize the objective function, ultimately selecting the shot with the highest combined immediate and future value. 
We also implement a simple greedy baseline agent \Ag\ that computes shot angle $\alpha$ via trigonometric approximation (other shot parameters are sampled randomly) to pocket each target ball in its closest pocket. \Ag\ performs shot selection to maximize the number of potted target balls using  random selection among equivalent shots. \Assistant\ can use any agent that fits the definition of $\A(\X)$.

%% file: src/method.tex
\section{Interactive and Explainable Pool Assistant}
\label{sec:method}
We explain three important aspects of \Assistant. In Section~\ref{sec:method:event-based-representation} we introduce our simple scheme to represent simulation traces as sequences of events described textually. Then, in Section~\ref{sec:method:interactive} we define our notion of interactive and explainable agents and explain the three modules of \Assistant\ that communicate with a physics simulator and an LM. Finally, in Section~\ref{sec:uncertainty}, we explain how we train a simple neural network to behave as an empirical surrogate for any given agent and how this surrogate encodes uncertainty.

\subsection{Event-based representation of simulation traces}
\label{sec:method:event-based-representation}
To facilitate language-based reasoning, we instrumented the simulation as $\Simi: \mathcal X \times \Theta \rightarrow \mathcal X \times \mathcal S$ to output $\Ev \in \mathcal S$, an ordered sequence $\Ev = [ \Ev_0, \Ev_1, \Ev_2, \cdots ]$  of predetermined events where each event \Evi\ is a concatenation of two string tokens followed by information tokens:
\begin{enumerate} [leftmargin=1.1em,itemindent=.5em]
    \item \texttt{ball}-\texttt{ball}-<bid1>-<bid2> when ball bid1 collides with ball bid2;
    \item \texttt{ball}-\texttt{cushion}-<bid> when ball bid bounces off cushion; and
    \item \texttt{ball}-\texttt{pocket}-<bid>-<pid> when ball bid falls into pocket pid
\end{enumerate}
which serve as an abstraction of state information ignoring details such as geometry and dynamics attributes (velocity, spin, etc.). Although such an encoding of simulation traces is lossy, it enables qualitative comparisons of shots and their outcomes. 

\subsection{Interactivity and explainability}
\label{sec:method:interactive}
We introduce agents $\AL: \mathcal X \times \mathcal L \rightarrow \Theta \times \mathcal L$. Given a contextual reasoning  query $\Q$ in natural language $\mathcal L$ and state \X, each such agent $\AL(\X,Q)$ returns a shot \T\ along with an explanation $\E \in \mathcal L$ that is rooted in domain expert rules.
We design a family of such agents using three sequential stages: A \emph{recommender} hypothesizes shots relevant to a state of the table \X\ and contextual query \Q; a \emph{tuner} optimizes the recommended shots to obtain $\T^*$ as the shot that best addresses the contextual query; and finally an \emph{explainer} that generates explanations \E\ based on rules $\mathcal R$.  
The recommender first queries an out-of-the-box LM to generate potential sequences of events $ L = \LM (Q, C), \; L,Q,C \in \mathcal L$ where $Q$ combines the user query with a prompt requesting relevant event sequences and $C$ is a context specifying a description of state and a list of ball IDs for the agent to target. Chain-of-Thought prompting \cite{wei2022CoTPromptingElicitsReasoning} is utilised, to allow the recommender to plan the shots it will suggest before producing the event sequences. Details of this process are included in Appendix B.  
The central role of the recommender is to find $N_r\in\mathbb{N}$ hypothetical shots $\hat{\T}_k$ that result in a sequence of simulation events $L_k$ (via $\Simi(\X,\hat{\T}_k)$) that resembles $L$. We achieve this via minimization of 
$ 
d_{\Ev}(L, L_k) - \lambda(|L_k| + \|\hat{\T}_k\|)
$
using simulated annealing, where $d_{\Ev}(L, L_k)$ yields the longest common (ordered) subsequence of $L$ and $L_k$ that necessarily begins with the first event in $L$. The second term is used to penalise exaggerated values. Finally, each shot has a strategy label $s_k \in \{\text{offensive}, \text{defensive}, \text{none}\}$ and a difficulty label $d_k \in \{easy, medium, hard, none\}$ based on the shot's desired characteristics: $ (s_k, d_k) = \LM (Q, C, L_k)$. These labels are used in Section~\ref{sec:strategy}. \\
Given \X\ and $\{\hat{\T}_k\}$, the tuner first evaluates $\X_k=\Simi(\X, \hat{\T}_k)$ for each $k$ and then evaluates $\mathbf r^k$ for each $k$ using the expert rules so that $\mathbf r_i^k = r_i(\X_k, \hat{\T}_k), i=1,2,\cdots,m$. For some rules, this requires knowledge of the result of executing the shot which we obtain via $\Sim(\X,\hat{\T}_k)$. The tuner then iteratively optimizes $\hat{\T}_k$ until the expected value (explained in \ref{sec:uncertainty}) of the shot is maximized. The tuner returns the highest-value shot $\T^*$ as the shot selected by $\AL(\X,Q)$ along with $\mathbf r^*$, the corresponding vector of rule function evaluations. 
The explainer $e$ augments a skeleton prompt $P$ containing general information such as a description of the task, table specifications, text-encodings of all events and some example sequences with contextual information. The  contextual information $C_e$ includes the positions of balls in \X, the dictionary of shot parameters \T\ and a list of rules and their value functions evaluated on $(\X,\T)$. Each field is quantized and encoded as a string. The final explanation is generated by an LM as $\E=\LM(P_e)$ where the augmented prompt $P_e=e(P,C_e)$. We refer readers to Appendix B for details and examples of $P_e$ and $C_e$.

\subsection{Incorporating uncertainty via neural approximation}
\label{sec:uncertainty}
We build an empirical surrogate \AN\ for any agent \A\ by training it to approximate the distribution $\cp {v} {\X} {\T} {\A}$ of the expected win rate of a shot \T\ given a table state \X. Since our goal is to use \AN\ instead of \A\ within an explainable agent \AL, for explanations to be grounded in the set of expert rules $\mathcal R$, we  define \AN\ to be a mapping from $\mathbf r \in [0,1]^m$ (rather than from $(\X,\T)$). The elements of $\mathbf r$ are $r_i(\X,\T)$. We discretize $\cp {v} {\X} {\T} {\A}$ using $n$ bins and therefore $\AN: [0,1]^m \rightarrow [0,1]^n$. At test time, we use the evaluated $\vn = \AN(\mathbf r)$ to assess the quality of the agent's shot. A high expectation indicates a good payoff and a low variance implies robustness to imperfections in execution.

Our environment models execution errors as Gaussian noise $\epsilon \sim \mathcal{N}(0, \sigma^2)$ injected into each shot \T\ given a table state \X. The execution of the perturbed action results in a stochastic state variable $\X_1 = \Sim(\X, \A(\X)+\epsilon)$. The \emph{value} of \T\ is the expected win rate $v\in[0,1]$ for player 1, which may be estimated via Monte Carlo  average over $M$ sampled $\X_1$ and simulating $N$ complete games. Instead, we estimate the probability distribution $\cp {v} {\X} {\T} {\A}$ which encodes uncertainty surrounding the quality of the shot. We use a simple Monte Carlo method, that averages $n-$bin histograms $\mathbf p \in [0,1]^n$ of the $M$ expected win ratio $\mathbf v_j,\; j=1,2,\cdots, M$ (for each sampled $\X_1$). Ideally this would be a tight distribution with a peak at $v=1$.  We use these samples to train our neural adaptor $\AN: [0,1]^m \rightarrow [0,1]^n$ (see Algorithm~\ref{alg:training}), which consists of a simple multi-layer perceptron network (MLP) with six layers of size 256 and ReLU activation functions. We train this using a Cross-Entropy loss function with a batch size of $128$, a learning rate of $0.005$, a dropout rate of $0.25$ over $25$ epochs.

\subsection{Incorporating strategy and difficulty}
\label{sec:strategy}
\label{sec:objective_function}

While the strategy and difficulty labels $s$ and $d$ characterize a shot, we additionally define binary strategy vectors $\mathbf{w}_o, \mathbf{w}_d \in \{0, 1\}^m$ containing classification across rules (on \textit{offensive} vs \textit{defensive}). Then we use these to calculate the strategy score $v_s$ and the difficulty score $v_d$ for each shot. If $\mathbf r_t$ is the rule evaluations associated with a shot $(\X_t, \T_t)$ being tuned, and if $s$ is \textit{defensive}, then $v_s=(\mathbf{w}_d - \mathbf{w}_o) \cdot \mathbf{r}_t$. If $s$ is \textit{offensive} then $v_s=(\mathbf{w}_o - \mathbf{w}_d) \cdot \mathbf{r}_t$.
We calculate a difficulty score $v_d = H_\mathrm{max} - |h_t - h_d|$ where $h_t =H(\AN(\mathbf r_t))$ is the entropy of the approximate value distribution and $h_d$ is the mean entropy of the $d^{th}$ section of the distribution of entropies across all samples in the dataset-- we partition the entropy  distribution across training samples into $3$ difficulty sets: low $[0,0.4)$, medium $[0.4,0.8)$ or high.


The tuner in Section~\ref{sec:method:interactive} combines these terms with uncertainty to perform shot optimization, to obtain $\T^*_k$ by maximizing the objective function
$
    \mathbb{E}\left[ \vn \right] + v_s + v_d
$
 for each shot $\hat{\T}_k$.

\begin{algorithm}
    \caption{   \label{alg:training} Generating one training sample}
     \KwIn{ \A, \Sim, \X, $\sigma$ \tcp*[r]{{Agent, sim func., state, noise level} }\\
            \quad \,\, $M$, $N$, $m$, $n$ 
            \tcp*[r]{\#samps of $\cpp {v} {\X} {\T}$, \#sims, \#rules, \#bins} 
        }
    \KwOut{Training sample $(\mathbf r, \mathbf p)$}
            
    $\T \gets \A(\X)$\;
    $\mathbf r = $ vector containing $r_i(\X,\T), \; i=1,2,\cdots,m $\; 

    \For{$j \gets 1$ \KwTo $M$}  { 
        $\epsilon \gets \mathcal N (0,\sigma^2)$\;
        $\X_1 \gets \Sim(\X, \T + \epsilon)$\;
        
        \For{$k \gets 1$ \KwTo $N$}{
            $\X_{\mathrm{end}} \gets $ SimulateGame($\X_1$, \A, $\sigma$)\;
            $\mathbf w_k \gets $ DidPlayer1Win$(\X_{end})$ \tcp*[r]{binary result} 
        }
        $\mathbf v_j \gets \sum_{k=1}^N \mathbf w_k / N$ \;
     }
        $\mathbf p \gets $ Histogram$(\mathbf v, n )$\; 
     \Return $(\mathbf r, \mathbf p)$\;
    \end{algorithm}

\subsection{Summary}
Given an agent \A, we first train its neural surrogate \AN\ that can rapidly approximate the distribution of win-rates of \A\ as player 1 from any state \X. Then, we embed the surrogate within the tuner of \Assistant, resulting in agent \AL\ which can be queried and provides explanations in natural language. That is, for each agent \A\ we can associate its three variants as a tuple $(\A,\AN,\AL)$. We implemented comparisons for the PoolMaster agent $(\Apm,\ANpm,\ALpm)$ and for a simple greedy agent $(\Ag,\ANg,\ALg)$ as described in Section~\ref{sec:performance}.

%% file: src/experiments.tex
\section{Experiments}
\label{sec:experiments}

\subsection{Qualitative evaluation across diverse queries}
\label{sec:qual_interactivity}

We demonstrate the interactive capabilities of \Assistant\ through four qualitative examples of diverse queries in Figure~\ref{fig:qual_examples}.
Given an initial state $\X_0$ (shown in the center of the figure), each query $Q \in \mathcal{L}$ is executed as $(\T^*, E)=\AL(\X_0, Q)$ with $5$ candidate shots and $300$ optimization steps as hyperparameters. The recommender generates candidate shots which are optimized by the tuner to select $\T^*$, followed by explanation generation via the explainer component $E$. The results of this are presented in Figure~\ref{fig:qual_examples} to demonstrate the system's interactive shot generation and explanation capabilities.



\subsection{Quantitative evaluation of performance, explainability}

\subsubsection{Performance} 
\label{sec:performance}

We evaluate agent performance through pairwise games in our \OurPool\ environment of three versions each of two core agents-- PoolMaster (pm) and greedy (g). For each of these agents, we have the agent (\Apm, \Ag), their neural surrogates (\ANpm,\ANg) and their \Assistant\ versions (\ALpm, \ALg). For each pair of these 6 agents, we simulate  100 games with 50 games each as player 1 (to mitigate potential starting position bias). The mean percentage win rates over the 100 games (and standard deviations) are tabulated in Table~\ref{tab:winrates} with each cell showing the win rate of the agent on the row as player 1 against the agent on the column as player 2. Standard deviations were calculated by fitting binomial distributions.

The hyperparameters for each agent are defined as follows: For \Apm, the search grid size $G$ is set to [15, 30], with shot parameters $\Theta = \{v, \beta, a, b\}$ to be optimised with 5 discretized intervals each. For neural-based agents ($\AN$, $\AL$), we set the number of candidate shots $K=3$ and use $N=300$ simulated annealing steps. For $\AN$, the $K$ candidate shots begin as randomly sampled parameters, before being tuned. Importantly, for $\AL$, the query prompt provided to the agent is simply: "\textit{Find the best shot for me in this position}". The networks for each agent $\{ \ANpm,\ANg, \ALpm, \ALg \}$ are trained on 2500 state-shot pairs generated from the \Apm\ or \Ag\ agents, as described in Section~\ref{sec:uncertainty}.

\subsubsection{Relevance to expert rules}
\label{sec:expert_rules_validation}
The reliability of the explanations provided by \Assistant\ hinges on the ability of the LM to assess the relevance of each expert rule to a given state-shot pair. This is used by the explainer within its context $C_e$. We verified that the LM estimates relevances that match ground truth rule evaluations. The dataset for this is generated through a two-phase sampling process. First, we extract state-shot pairs $(\X,\T)$ from the network training dataset and apply K-means clustering ($K=25$) to their rule evaluation vectors, then sample uniformly from each cluster to create a diverse evaluation set that captures varying shot characteristics. State-shot pairs $(\X_r,\T_r)$ for evaluation are sampled from this dataset, evaluating the ground truth $r_i(\X,\T), \, i=1,2.\cdots,m$ and classifying them according to a Likert scale quantizing the $[0,1]$ range of $r_i$ into bins:\\
\begin{tabular}{>{\centering\arraybackslash}m{0.75cm} >{\centering\arraybackslash}m{0.75cm} >{\centering\arraybackslash}m{0.75cm} >{\centering\arraybackslash}m{1.5cm} >{\centering\arraybackslash}m{0.75cm} >{\centering\arraybackslash}m{0.75cm} >{\centering\arraybackslash}m{0.75cm}}
    \cellcolor{gray!5} 0 & \cellcolor{gray!10} 1 & \cellcolor{gray!15} 2 & \cellcolor{gray!20} 3 & \cellcolor{gray!25} 4 & \cellcolor{gray!30} 5 & \cellcolor{gray!35} 6 \\
    \cellcolor{gray!5} very low & \cellcolor{gray!10} low & \cellcolor{gray!15} mod low & \cellcolor{gray!20} moderate & \cellcolor{gray!25} mod high & \cellcolor{gray!30} high & \cellcolor{gray!35} very high \\
\end{tabular}\\
all bins are $0.125$ wide except for `moderate' which is twice as wide and spans the interval $[0.375,0.625)$. Each element in our reference dataset consists of $(\X_r, \T_r, \mathbf r_r, \mathbf s_r, \mathbf K_r)$ where $\mathbf r_r$ is an $m-$vector vector containing all reference evaluations $r_i(\X,\T)$,  $\mathbf s_r$ is an $m-$vector of classified bin numbers for each rule  and $K_r$ is an $m-$vector of strings containing the corresponding Likert keys. 

Our experiment constructs $C_e^r$ from $(\X_r, \T_r, \mathbf r_r)$ (as done by the explainer described in Sec.~\ref{sec:method:interactive}) and uses this to augment a skeleton test prompt $P_t$. This prompt queries the LM to predict the vector of Likert values $\mathbf s_t$ by comparing the keys $K_t$ given the context $C_e^r$. 
Further details are presented in Appendix C. 
Ideally, the distance $|\mathbf s_t - \mathbf s_r|$ between the test and reference values should be $<1$. Figure~\ref{fig:exp2_70B} plots the mean distances (orange bars) for each rule (X-axis) along with error bars across the dataset. The plot on the right side shows the aggregate over all rules. Figure~\ref{fig:exp2_70B} also shows distances (blue bars) from a control method where the context is constructed from only the state-shot pair $(\X_r, \T_r)$ instead of  $(\X_r, \T_r, \mathbf r_r)$.

\subsubsection{User study} 
\label{sec:user_study}
We evaluate explanation quality through a comparative user study between \Assistant\ and a baseline LM. Our participant pool $\mathcal{U}$ consists of 100 users recruited through Prolific \cite{prolific}. Each participant evaluated 20 unique state-shot pairs sampled from the diverse dataset described in Section~\ref{sec:expert_rules_validation}.
Each evaluation instance presents participants with an animated GIF visualization of the state-shot pair $(\X_i, \T_i)$ with simulated frames, an explanation generated by either \Assistant\ or the baseline model, and a 7-point Likert scale for quality assessment. Participants provide ratings $r_{u,i} \in \{1,...,7\}$ for each instance, with a mandatory 10-second viewing period before rating is allowed. 
Examples of the user prompt are presented in Appendix E.
The baseline model follows the same explanation generation protocol as \Assistant\ (Section~\ref{sec:method:interactive}) but without access to rule function evaluations $\mathbf{r}$, while retaining access to the natural language rule descriptions ${R_i}$. Post-study, participants self-report their pool expertise $l_u \in \{\text{None}, \text{Low}, \text{High}\}$.
Figure~\ref{fig:exp3_distributions} presents the rating distributions seperated by expertise level.




%% file: src/table.tex
\begin{table}
    \begin{tabular}{@{}>{\columncolor{gray!30}[0.1em]}l@{\,\,\,}cccccc@{}}
    \rowcolor{gray!30} 
    \cellcolor{white} & \ANpm & \ANg & \ALpm & \ALg  & \Apm  & \Ag 
    \\
    \ANpm & - & \textbf{81 {\color{gray!85}(3.9)}} & \textbf{73 {\color{gray!85}(4.4)}} & \textbf{77 {\color{gray!85}(4.2)}} & \textbf{62 {\color{gray!85}(4.8)}} & \textbf{86 {\color{gray!85}(3.5)}} 
    \\
    \ANg & 19 {\color{gray!85}(3.9)} & - & 55 {\color{gray!85}(5.0)} & 76 {\color{gray!85}(4.3)} & 35 {\color{gray!85}(4.8)} & 75 {\color{gray!85}(4.3)} 
    \\
    \ALpm & 27 {\color{gray!85}(4.4)} & 45 {\color{gray!85}(5.0)} & - & 64 {\color{gray!85}(4.8)} & 59 {\color{gray!85}(4.9)} & 77 {\color{gray!85}(4.2)} 
    \\
    \ALg  & 23 {\color{gray!85}(4.2)} & 24 {\color{gray!85}(4.3)} & 36 {\color{gray!85}(4.8)} & - & 27 {\color{gray!85}(4.4)} & 67 {\color{gray!85}(4.7)} 
    \\
    \Apm  & 38 {\color{gray!85}(4.8)} & 65 {\color{gray!85}(4.8)} & 41 {\color{gray!85}(4.9)} & 73 {\color{gray!85}(4.4)} & - & 85 {\color{gray!85}(3.6)} 
     \\
    \Ag  & 14 {\color{gray!85}(3.5)} & 25 {\color{gray!85}(4.3)} & 23 {\color{gray!85}(4.2)} & 33 {\color{gray!85}(4.7)} & 15 {\color{gray!85}(3.6)} & - 
    \\
    \end{tabular}
    \caption{Win rates (\%) between agents. The percentage win rate (std. deviations) of Player 1 agents (row) against Player 2 agents (columns) are tabulated. The agents are: surrogate PoolMaster (\ANpm), surrogate greedy (\ANg), CueTip PoolMaster (\ALpm), CueTip greedy (\ALg), PoolMaster (\Apm) and greedy agent (\Ag). }
    \label{tab:winrates}
\end{table}

%% file: src/results.tex
\section{Results}
\label{sec:results}

\begin{figure*}
    \centering
    \includegraphics[width=.9\textwidth]{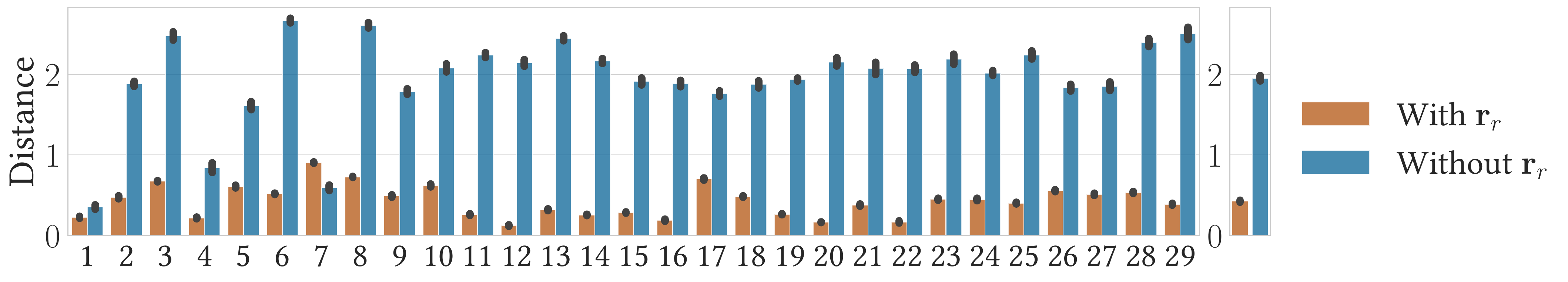}
    \caption{
    \textbf{Left:} Distributions of Likert scale distances between ground truth and estimations from different methods for each domain expert rule (mean$\pm$std. error) ; \textbf{Right:} Aggregated results over all domain expert rules (mean$\pm$std. error) ; Results obtained using the LM \textit{Llama-3.1-70B-Instruct}~\citep{llama3modelcard,llama-3-70B}.}
    \label{fig:exp2_70B}
\end{figure*}

\begin{figure*}
    \includegraphics[width=0.9\columnwidth]{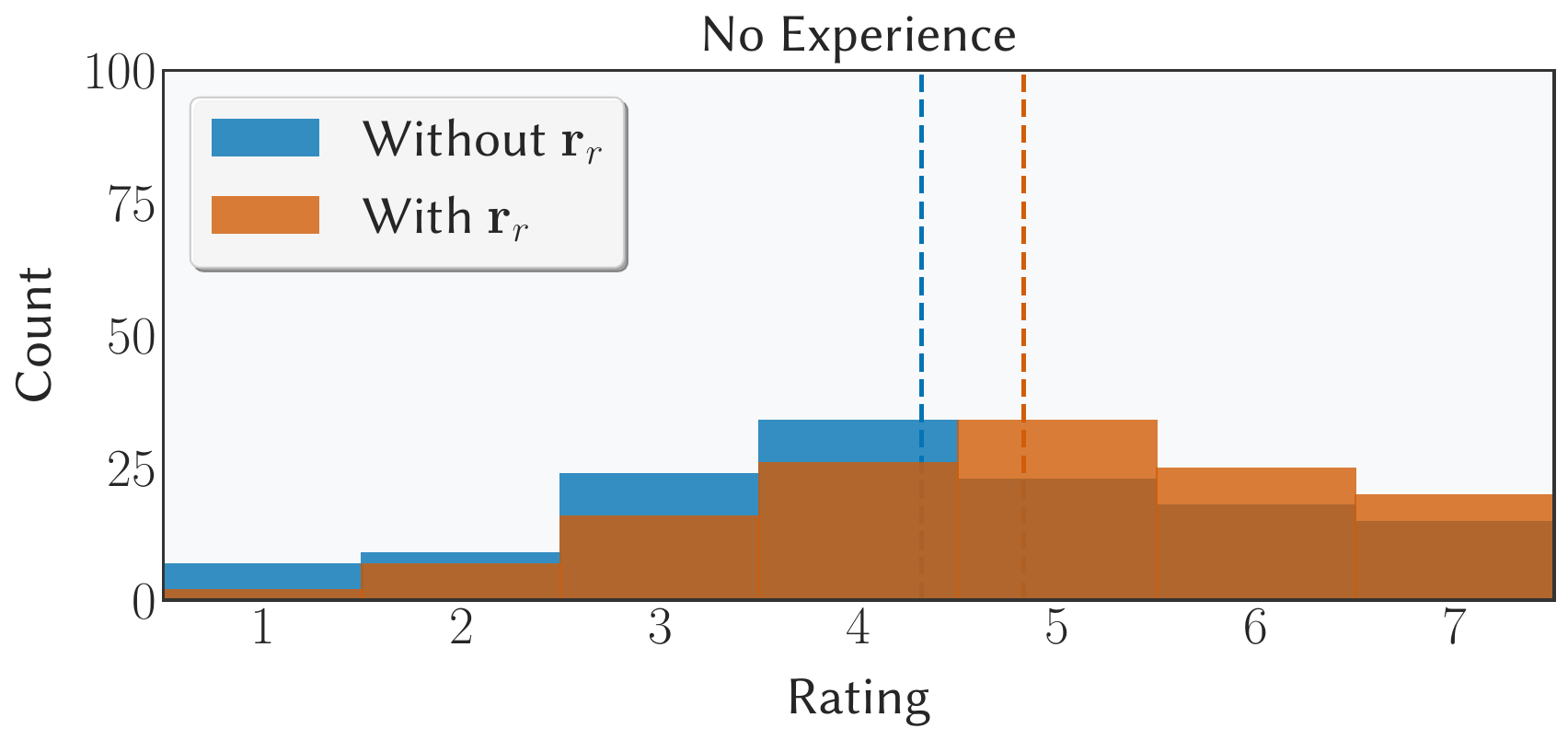}
    \includegraphics[width=0.9\columnwidth]{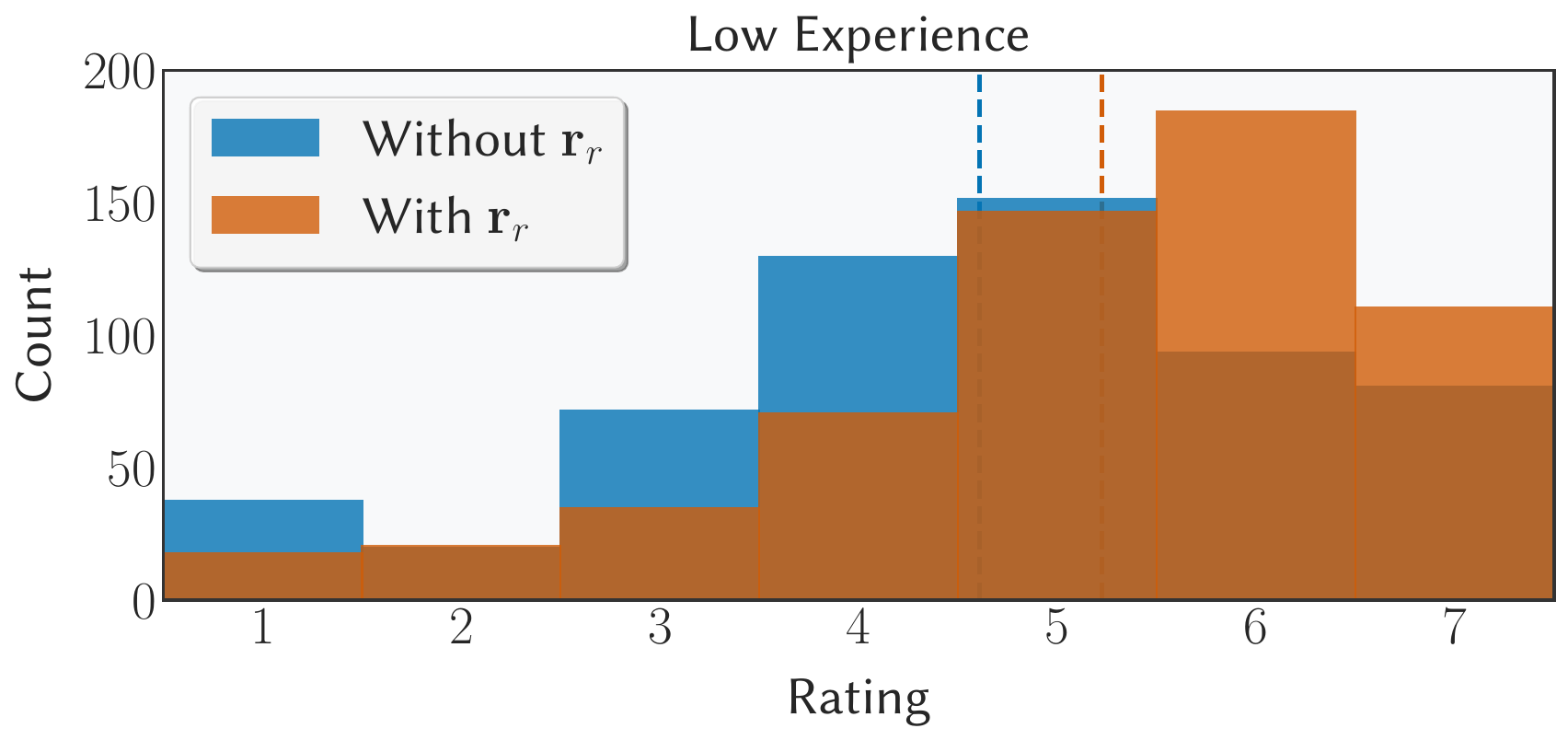}
    \includegraphics[width=0.9\columnwidth]{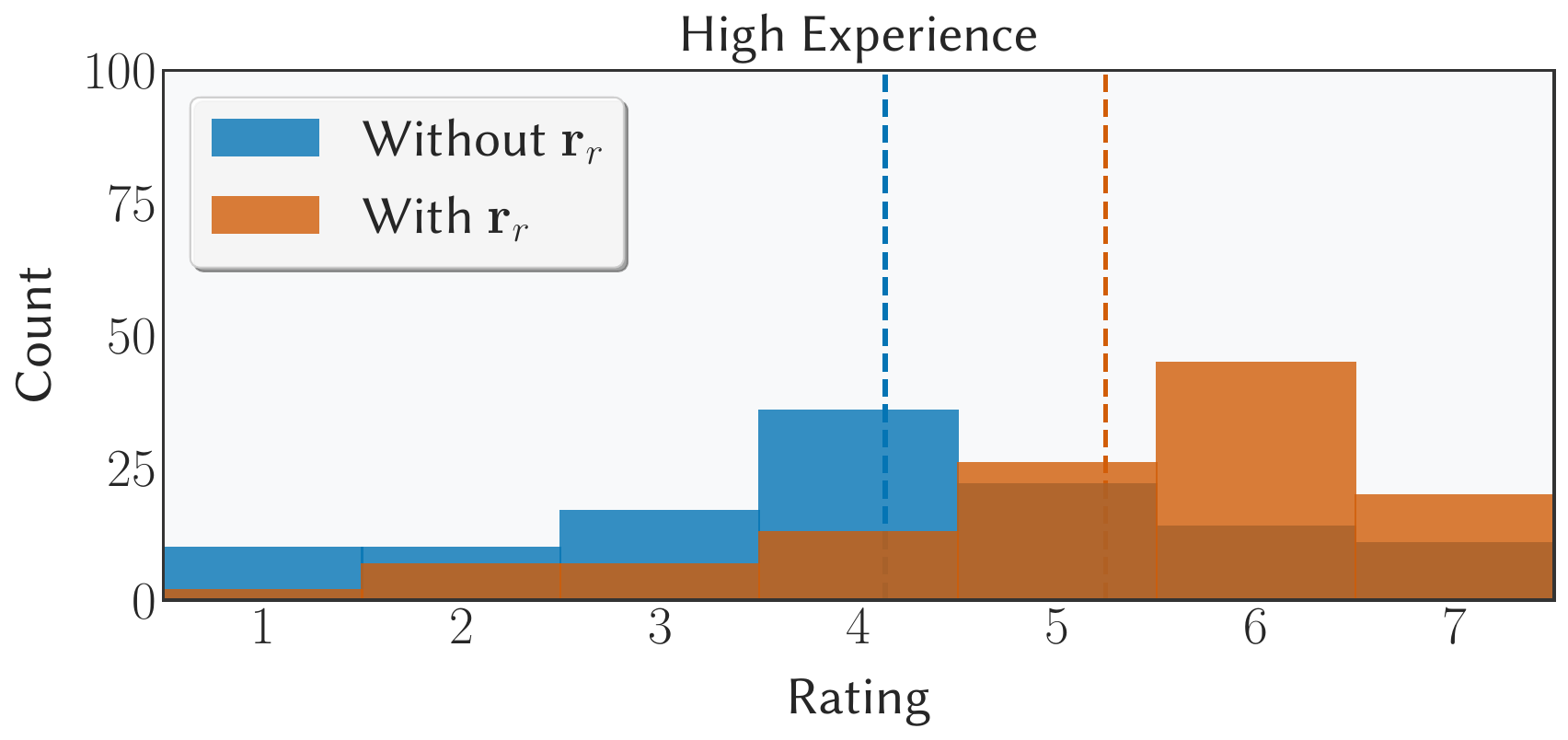}
    \caption{Distributions of ratings given by users in our survey, where each user self-reported their expertise. Users who self-reported as having high experience (bottom) rated \Assistant's explanations higher than the baseline.}
    \label{fig:exp3_distributions}
\end{figure*}

\subsection{Qualitative evaluation of our agent}


Figure~\ref{fig:qual_examples} presents four queries given to \Assistant. The generated shots are seen to be tailored to the query in each case. For example, given the query "I am feeling defensive, find me a defensive shot", \Assistant\ generates event sequences with strategy label $s=\text{defensive}$ and difficulty label $d=\text{medium}$ (not illustrated). The tuning process, guided by these labels, produces a shot $\T^*$ that maximizes defensive positioning, as evidenced in the motion trajectory visualization. The explainer component grounds this shot in expert rules ${R_i}$, where defensive rule $R_5$ (safety opportunity identification) applies positively (hence shown in green) while offensive rule $R_{13}$ (target ball potting prioritization) does not apply (hence shown red), demonstrating the effectiveness of the strategy-aware optimization.

For "Show me a trick shot!", \Assistant\ demonstrates its capacity to generate a high-difficulty shots by setting $s=\text{offensive}$ and $d=\text{hard}$. The selected shot $\T^*$ exhibits multiple complex characteristics: a long distance to travel and multiple cushion collisions. The neural adaptor $\AN$  predicts an expected value of $0.34$, indicating significant execution difficulty (not illustrated). The explainer identifies relevant rules including $R_{14}$ (distance-based difficulty scaling) and $R_{29}$ (multi-cushion complexity), reflecting its high-risk nature.

\subsection{Quantitative evaluation of our agent}
\subsubsection{Performance}
The win rates in Table~\ref{tab:winrates} show that the neural surrogate \ANpm\ performas best, achieving the highest win rate. Notably, \ANpm\ achieves a win rate exceeding 50\% against its training baseline $\Apm$ (winner of the computational pool tournament). Results suggest that the neural adaptor's shot optimization strategy (with stochasticity and uncertainty) surpasses the agents used to train the adaptor.
The greedy surrogate, \ANg\ , exhibits reduced performance with a 35\% win rate against \Apm, highlighting the importance of the underlying agent used during training. The interactive agents \ALpm\ and \ALg\ maintain strong performance, with \ALpm\ achieving a 59\% win rate against $\Apm$.
The interactive agents \ALpm\ and \ALg\ exhibit a consistent performance reduction compared to their non-interactive counterparts, highlighting the trade-off between performance and interactivity.

\subsubsection{Relevance to expert rules} 
\label{sec:exp2-results}
Figure~\ref{fig:exp2_70B} plots the results of the experiment presented in Section~\ref{sec:expert_rules_validation}. 
Since the error is $<1$ on average, for the orange bars, the experiment verifies that a combination of textual Likert keys and rule evaluation vector $\mathbf r_r$ ensures that the LM's evaluation of the current state is aligned with the reference.
The blue bars represent a higher error (p-values $<1e-4$ in a two-sample Kolmogorov Smirnov test, except for rule $7$). Thus $\mathbf r_r$ plays a vital role in grounding the explanations in the expert rules. 

\subsubsection{User study} 
Figure~\ref{fig:exp3_distributions} presents the rating distributions, seperated by user expertise, from our experiment described in Section~\ref{sec:user_study}. The distributions show that \Assistant-generated explanations consistently achieve higher mean ratings across all expertise levels $l_u \in \{\text{None}, \text{Low}, \text{High}\}$.
For users with no prior pool experience, we observe a moderate difference in mean ratings ($\mu_{\Assistant} = 4.83$, $\mu_{\text{baseline}} = 4.32$, $\Delta\mu = 0.51$). This difference increases slightly for users with low experience ($\mu_{\Assistant} = 5.22$, $\mu_{\text{baseline}} = 4.61$, $\Delta\mu = 0.61$), and is most pronounced among highly experienced users ($\mu_{\Assistant} = 5.24$, $\mu_{\text{baseline}} = 4.13$, $\Delta\mu = 1.11$). 
The rating differential in the \textit{High Experience} group suggests highly experienced players believe that \Assistant's explanations better reflect true shot characteristics. We attribute this to \Assistant's access to the expert rule evaluations $\mathbf{r}$, which grounds explanations in accurate shot information rather than relying on a LM's interpretation of the expert rules ${R_i}$.

%% file: src/conclusion.tex
\section{Implementation and discussion}
\paragraph{Winrates and Potting Rates}
Analysis of the Section~\ref{sec:performance} experiments reveals an important distinction between win rates and potting rates (defined as the proportion of shots that successfully pot a target ball). The baseline \Ag\ achieves the lowest potting rate at 29\%, consistent with its poor performance, while \Apm\ demonstrates strong performance with 82\%. \ALpm\ achieves the highest potting rate at 93\%, indicating highly effective shot recommendation and optimization.
Notably, \ANpm\ achieves the highest win rate despite a relatively low potting rate of 62\%, suggesting a more nuanced strategy that balances offensive and defensive play. This demonstrates that optimal performance in pool requires strategic shot selection beyond simple potting optimization, a capability effectively captured by the neural surrogate approach.

\paragraph{LM impact on reliability}
We employ zero-shot chain-of-thought (CoT) reasoning with majority-voting self-consistency \citep{wei2023chainofthoughtpromptingelicitsreasoning,wang2023selfconsistencyimproveschainthought} for LM queries. Our analysis spans the Llama 3 model family~\cite{llama3modelcard}, comparing \textit{Llama-3.1-3B-Instruct}, \textit{Llama-3.1-8B-Instruct}, and \textit{Llama-3.1-70B-Instruct} variants in their ability to assess expert rule relevance for state-shot pairs.
Performance scales positively with model size, as evidenced by decreasing Likert distance between model estimations and expert rule evaluations: $1.99\pm0.40$ vs $2.40\pm0.45$ for $3$B, respectively, with $\mathbf r_r$ vs without $\mathbf r_r$ ; $1.49\pm0.61$ vs $2.34\pm0.51$ for $8$B, respectively, with $\mathbf r_r$ vs without $\mathbf r_r$ ; and $0.43\pm0.19$ vs $1.95\pm0.55$ for $70$B, respectively, with $\mathbf r_r$ vs without $\mathbf r_r$. 
Additional experiments with prompt optimization using DSPy~\cite{khattab2024dspy} show statistically significant improvements only for the medium-sized model. 
Based on these results, we employ the $70$B parameter model for all experiments presented in this work. 
Details about the prompts and detailed scaling analysis are provided in Appendix C.

\paragraph{Agent ablation}
We conduct an ablation study to quantify the impact of the neural surrogate in the \Assistant\ system by defining two variants: $\AL_{\text{baseline}}$ and $\AL_{\text{h}}$. Both maintain the recommender component but replace the neural tuner with base LM decision-making and heuristic-guided LM decision-making respectively.
The average win rates against all agents are: $\AL_{\text{baseline}}$ (26\%), $\AL_{\text{h}}$ (32\%), \Ag\ (22\%), \ANpm\ (75.8\%), and \ANg\ (52\%), \ALpm\ (54.4\%), \ALg\ (37.4\%), and \Apm\ (60.4\%). While the heuristic guidance of $\AL_{\text{h}}$ improves LM decision-making over the baseline $\AL_{\text{baseline}}$, the substantial performance gap between LM and neural-based variants ($\Delta > 40\%$) demonstrates the neural surrogate's critical role in \Assistant's performance.

\paragraph{Expert rule significance}
We analyze rule significance using the diverse evaluation dataset from Section~\ref{sec:expert_rules_validation} by computing correlations between \ANpm's potting rates and expert rule evaluations for each state. The rules exhibiting strongest negative correlation with potting success are $R_{16}$ (obstacle presence), $R_{28}$ (multi-ball collisions), and $R_{29}$ (multi-cushion interactions). These correlations reveal that shot complexity is most significantly impacted by physical obstruction and multiple collision events, aligning with the fundamental physics constraints of pool. 

\paragraph{Diversity of queries}
One of the benefits of a natural language is the scope for rich queries. 
The space of effective prompts is bounded by the search parameters defined, which in our case are: events (ball-ball collisions, ball-pocket, ball-cushion bounces), strategy (offensive or defensive), and difficulty (easy, medium, hard). As long as the query pertains to tactics that operate within this space, they may be arbitrarily formed. To enable queries outside this space--for example, ``make the ball bounce off the left cushion''-- would require introducing a special event \textit{ball-cushion-left}. However, this addition would require no new training, since the network surrogate depends on the actual simulation state rather than the event sequences and the LM is used off-the-shelf.
There is a trade-off between introduction of events to enrich the space of possible queries and speed of execution (larger search space for the recommender).



\paragraph{Upgrading the underlying agent}
\Assistant\ abstracts away the underlying agent that is responsible for tactical shot decisions via the neural surrogate discussed in Section~\ref{sec:uncertainty}. However, our results confirm that the quality of the underlying agent used to train the neural surrogate  impacts \Assistant's win-rate. To upgrade the underlying agent, the MLP within neural surrogate will need to be retrained using Algorithm~\ref{alg:training} where data is collected using the new agent.

\paragraph{Sharing the \OurPool\ environment}
While effective for playing \OurPool, some of the choices we make are domain-specific (such as the choice of important events) which limits potential claims about generalizability to other physical systems. However, we will share our implementation of \Assistant\ along with the \OurPool\ environment as an open-source benchmark. We hope that the environment will serve as a sandbox to test and inspire work on explainable agents in physics-based environments. The physics of \OurPool\ provide sufficiently rich physical dynamics (collision, friction, spin, finite element modelling of bounces off soft cushions, etc.) while being simple enough to allow for rapid prototyping, experimentation and visualization. 


\paragraph{Limitations}

Domain rules improve explainability at the cost of needing the rule evaluation functions $r_i$ to be implemented. Potential inconsistency between $R_i$ and the implementation of $r_i$ could introduce  inaccuracies in rule evaluation. The reliance on simulated annealing for shot optimization provides no guarantees of (global) optimality, although empirical results demonstrate robust performance.
Finally, the quality of our results is affected adversely by scaling down the LM which might be considered a practical limitation for some devices.

\section{Conclusion and future work}

Our experimental results demonstrate the effectiveness of utilising expert-provided heuristics to create interactive, explainable agents operating in a specific physics-based environments. The quantitative improvements in win rates, reliability, and explanation quality, in addition to the qualitative examples, show that grounding agent decisions in expert-provided heuristics, and replacing traditional agents with neural surrogates, enhances performance, interpretability, and interactivity.

This work suggests several promising research directions. First, the system could be expanded to enable automated discovery of heuristics through iterative optimization and environment feedback, potentially identifying novel domain insights while maintaining reliability and explainability. Second, future work could explore synthesizing unified domain-specific languages that better capture both action spaces and domain rules, enhancing the expressiveness of heuristics and the interpretability of agents. Both of these possibilities would make use of an iterative improvement mechanism using LMs, where a system can adapt and learn from online, physically simulated environments. Finally, the system could be extended to other physical reasoning tasks, such as robotics, where the ability to ground decisions in expert-provided rules could significantly enhance the interpretability and reliability of autonomous agents.

%% file: src/acknowledgments.tex
\section{Acknowledgments}

This work was funded by the UKRI Centre for Doctoral Training in Natural Language Processing (grant EP/S022481/1) and ELIAI (The Edinburgh Laboratory for Integrated Artificial Intelligence) EPSRC (grant EP/W002876/1). 
This work was conducted while the primary author was an NLP research intern at Tencent, London.

%% file: src/image_pages.tex
\begin{figure*}
    \centering  
    \includegraphics[width=.95\textwidth]{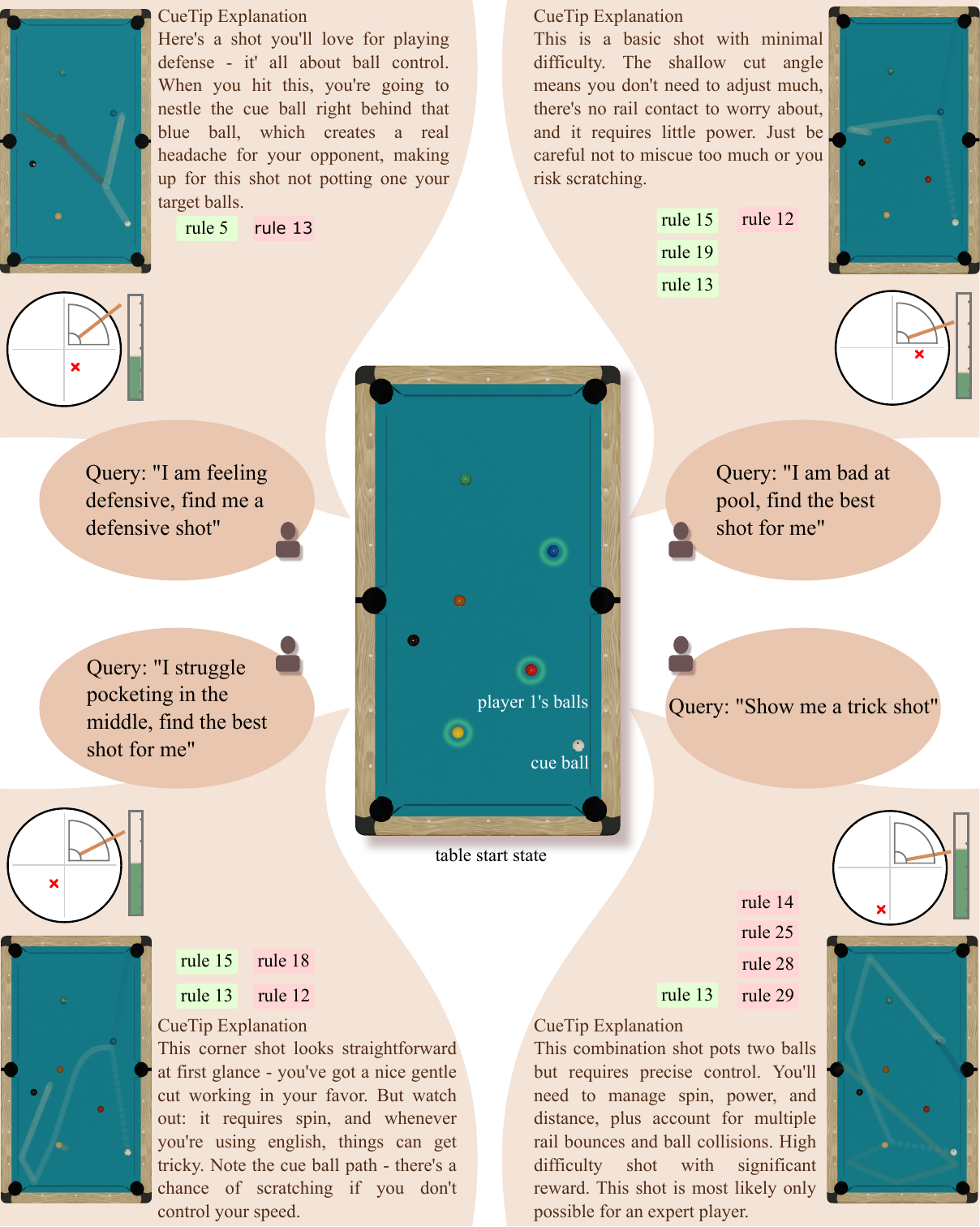}
    \caption{\label{fig:qual_examples}  The figure shows an initial state (center) where  player 1 needs to choose a shot. The player interacts with  \Assistant\ prompting it with the state of the table and a query. Four example queries are shown in different quadrants. \Assistant\ returns shot parameters (illustrated with a red `x' for point of contact, a brown line for angle from vertical and a green bar on the right for power) and an explanation. Also shown are rules from the expert guidelines (detailed in Appendix A) and whether they apply positively (green) or negatively (red). }
\end{figure*}

%% file: src/appendix.tex
\appendix

\section{Natural Language Description of the Domain Expert Rules}
\label{app:defs}


We detail in the following the natural language descriptions $R_i\in\mathcal L$ of the domain expert rules introduced in Section $2.2$:
\begin{enumerate}
\item Ball Groupings: Identify sets of two or more balls of the same type in close proximity that can be easily pocketed in sequence. These groupings increase the value of the table state as they allow for multiple shots without significant cue ball movement.
\item Makable Regions: Assess areas on the table where balls can be pocketed without using kick or bank shots. Pay special attention to overlapping makable regions for multiple balls, as these areas offer the most versatility and shot options.
\item Insurance Balls: Locate balls that can be easily pocketed from almost anywhere on the table. These serve as valuable backup options when positioning goes awry or when faced with a difficult layout.
\item Break-up Opportunities: Evaluate clusters of balls that need separation. Shots that can break up these clusters while pocketing a ball or achieving good position are particularly valuable.
\item Safety Opportunities: Identify chances to play defensive shots that leave the opponent in a difficult position. Good safety opportunities can be as valuable as offensive shots in many situations.
\item Two-way Shot Possibilities: Look for shots that offer both offensive and defensive potential. These shots allow for pocketing a ball while also setting up a good defensive position if missed, providing strategic flexibility.
\item Table Layout for Safeties: Assess the overall layout for defensive play. A valuable table state often includes options for effective safety plays if offensive shots become too risky.
\item Multiple-ball Positions: Consider the arrangement of multiple balls that need to be pocketed in sequence. A valuable layout allows for natural progression from one ball to the next without difficult positional play.
\item Avoidance Shots: Recognize balls that should be avoided to maintain a favourable layout or to leave the opponent in a difficult position. The ability to navigate around these balls adds value to the current state.
\item Combination and Bank Shot Opportunities: While often more difficult, the presence of viable combination or bank shots can add value to a table state by providing additional options.
\item{
Rail Proximity: Consider the position of balls near rails. While sometimes challenging, balls near rails can offer unique offensive or defensive opportunities.
}
\item Scratch Potential: Evaluate the layout for potential scratches. A valuable table state minimizes the risk of scratch shots while maximizing scoring opportunities.
\item Above all, prioritise shots that pot the most target balls.
\item Distance: Shot difficulty increases with greater distances between the cue ball, object ball, and pocket. Longer shots require more precise aim and speed control.
\item Cut Angle: Larger cut angles are more challenging than straight or small angle shots. The margin for error decreases as the cut angle increases.
\item Obstacle Balls: The presence of other balls obstructing the path of the cue ball or object ball significantly increases shot difficulty. This may require more precise positioning or the use of advanced techniques.
\item Rail Contact: Shots requiring the cue ball to hit a rail first (like rail cut shots) are more complex due to the need to account for rail dynamics and potential throw effects.
\item English Requirements: Shots needing side spin (English) are more difficult to control and execute. The use of English introduces additional variables that affect both aim and cue ball behaviour after contact.
\item Speed Control: Shots requiring precise speed control, whether very fast or very slow, are more challenging. Speed affects pocket geometry, throw, and positioning for the next shot.
\item Follow/Draw Needs: Shots requiring significant follow or draw are more difficult than natural roll shots. These shots demand precise vertical axis control of the cue ball.
\item Rail Proximity: Balls very close to rails can be more difficult to hit cleanly and may require specialized techniques like rail cut shots or spin.
\item Scratch Potential: Positions with a high risk of scratching are more difficult to play safely and effectively.
\item Spin Shots: Difficulty increases with the amount of curve required. These shots demand precise control of both vertical and horizontal spin.
\item Frozen Ball Situations: Balls touching each other or touching a rail create unique challenges, often requiring precise speed and spin control.
\item Multiple Effects: Shots involving a combination of factors (e.g., cut angle, speed, and English) are particularly challenging due to the need to account for multiple variables simultaneously.
\item Throw Effects: Accounting for throw (both cut-induced and English-induced) adds complexity, especially on longer shots or those with significant cut angles.
\item Deflection and Cue Ball Curve: When using English, especially at higher speeds or with an elevated cue, accounting for cue ball deflection and curve increases shot difficulty.
\item Multi-ball Collision: It is exponentially difficult to pot a ball by colliding it with multiple balls.
\item Multi-cushion Collision: It is exponentially difficult to pot a ball by having it bounce off multiple cushions.
\end{enumerate}

\newpage
\section{Language Model Prompts \& Implementation Details}
\label{app:lm-prompts}

The following sections detail the prompts used to query the LM as part of the \textit{recommender} (Section~\ref{app:lm-prompts:recommender}) and the \textit{explainer} (Section~\ref{app:lm-prompts:explainer}), based on the template prompt detailed Section~\ref{app:lm-prompts:template}.

For implementation purposes, we categorise the domain expert rules into value-related rules and difficulty-related rules.
The value-related rules consists of the first $13$ rules $\{R_i, r_i | i=1,2,\cdots,13\}\subset\mathcal R$ and they refer to concepts that contribute to the value of a shot.
The difficulty-related rules consist of the last $16$ rules $\{R_i, r_i | i=14,15,\cdots,29\}\subset\mathcal R$ and they refer to concepts that contribute to the difficulty of the shot.
We will refer to them, respectively, as $\mathcal R^v$ and $\mathcal R^d$.

\subsection{Template Prompt}
\label{app:lm-prompts:template}

We present a template prompt in Listing~\ref{lst:template} with interlocutors highlighted in blue with framing, and variables highlighted in green.
It corresponds mostly to the common template implemented by DSPy~\citep{khattab2024dspy} to reproduce (zero-shot) Chain-of-Thought prompting~\citep{wei2022CoTPromptingElicitsReasoning}, as seen in the \pgvar{CoT\_reasoning} variable.

\lstset{caption={Template Prompt}, label={lst:template}}
\begin{lstlisting}[escapechar=@]
@\pfVar{blue}{SYSTEM:}@
Your input fields are:
@\pgvar{list\_input\_fields}@

Your output fields are:
@\pgvar{CoT\_reasoning}@
@\pgvar{list\_output\_fields}@

All interactions will be structured in the following way, with the appropriate values filled in.

[[ ## @\pgvar{input\_field1.name}@ ## ]]
{@\pgvar{input\_field1.name}@}

@  $\cdots$ @
 
[[ ## @\pgvar{input\_fieldN.name}@ ## ]]
{@\pgvar{input\_fieldN.name}@}

[[ ## completed ## ]]

In adhering to this structure, your objective is: 
@ \pgvar{task\_description} @

@\pfVar{blue}{USER:}@
[[ ## @\pgvar{input\_field1.name}@ ## ]]
{@\pgvar{input\_field1.value}@}

@ $\cdots$ @
 
[[ ## @\pgvar{input\_fieldN.name}@ ## ]]
{@\pgvar{input\_fieldN.value}@}

Respond with the corresponding output fields, starting with the field `[[ ## @\pgvar{CoT\_reasoning}@ ## ]]`, then `[[ ## @\pgvar{output\_field1.name}@ ## ]]`, then @$\cdots$@, then `[[ ## @\pgvar{output\_fieldM.name}@ ## ]]`, and then ending with the marker for `[[ ## completed ## ]]`.

\end{lstlisting}

\subsection{Recommender Prompts}
\label{app:lm-prompts:recommender}

Recall that the role of the recommender is to find $N_r\in\mathbb{N}$ hypothetical shots such that $\hat{\T}_k$ that result in a train of events $\hat{L}_k$, produced by simulating $\Simi(\X,\hat{\T}_k)$, is similar to a desired/target train of events $L_k$, for each $k=1,2,\cdots,N_r$. 
The recommender obtains the target/desired train of events by querying an out-of-the-box language model for potential sequence of events that are in touch with the user's query $Q\in\mathcal L$ and the current context $C\in\mathcal L$, specifying a description of state and a list of ball IDs for the agent to target, as follows : $s_k, d_k, L_k = \LM (Q, C)$ where $s_k \in \{\text{offensive}, \text{defensive}, \text{none}\}$ and $d_k \in \{easy, medium, hard, none\}$ are, respectively, the strategy and difficulty terms.
The recommender leverages each train of events $L_k$ to find shot parameters $\hat{\T}_k$ via minimization of 
$ 
d_{\Ev}(L_k, \hat{L}_k) - \lambda(|\hat{L}_k| + \|\hat{\T}_k\|)
$
using simulated annealing, where $d_{\Ev}(L_k, \hat{L}_k)$ yields the longest common (ordered) subsequence of $L_k$ and $\hat{L}_k$ that necessarily begins with the first event in $L_k$. 
The output of the recommender is then a set of $N_r$ shots $\{\hat{\T}_k, s_k, d_k\}$ resulting from different initialization. \\

Thus, the recommender prompts the LM with the task description detailed in Listing~\ref{lst:recommender:td}, and the following inputs:
\begin{itemize}
\item{Balls information of the current state $\X\in\mathcal X$: the IDs and exact (x,y) coordinates of each ball currently on the table ;}
\item{Target balls information of the current state $\X\in\mathcal X$: the IDs of the balls that must be pocketed to win, and of the ones that must be avoided ;}
\item{Message $Q\in\mathcal L$: a message from a user to inform the decision-making process on what shots to suggest ;}
\item{Number of shots $N_r\in\mathbb{N}$: the number of shots to return;}
\end{itemize}

The list of outputs that it queries from the LM are as follows:
\begin{itemize}
	\item{Response: the strategy $s_k\in \{\text{offensive}, \text{defensive}, \text{none}\}$, the difficulty term $d_k \in \{easy, medium, hard, none\}$, and the train of events $L_k\in\mathcal L$ for each shot to suggest to the user, for $k=1,2,\cdots,N_r$.
}
\end{itemize}

\lstset{caption={Recommender Prompt - Task Description}, label={lst:recommender:td}}
\begin{lstlisting}[escapechar=@]
You are tasked with suggesting shots to make in a game of pool/billiards. Based on the message you recieve and the current state of the pool table, you must suggest shots to make that satisfy the users goal. 

The IDs of the pockets are: left top (lt) at (0,2), right top (rt) at (1,2), left center (lc) at (0,1), right center (rc) at (1,1), left bottom (lb) at (0,0), right bottom (rb) at (1,0).

The description of a shot is created using the events that you wish to occur, using the notation: "BALL-BALL-X-Y" for a collision between two balls X and Y, "BALL-POCKET-X-Z" for a ball X falling into pocket Z, and "BALL-CUSHION-X" for a ball X colliding with a cushion (note there is no second argument needed). Output the events of each shot as a comma separated list, with each shot on a new numbered line, for example:
1. BALL-BALL-cue-blue, BALL-CUSHION-blue,BALL-POCKET-blue-lb
2. BALL-BALL-cue-red, BALL-BALL-red-blue, BALL-POCKET-red-rc 
3. ... 
N. BALL-CUSHION-cue, BALL-CUSHION-cue, BALL-BALL-cue-yellow, BALL-POCKET-yellow-lt

An example of the reasoning to perform:
```
The blue ball is near the rc pocket, so we should aim for the rc pocket. But it looks like the red ball is between the cue ball and blue ball based on the following coordinates ... We'll need to bounce the cue ball off of a cushion ...
```

and an example response:
```
STRATEGY: offensive
DIFFICULTY: easy
SHOTS:
1. BALL-BALL-cue-blue, BALL-CUSHION-blue, BALL-POCKET-blue-rc
2. BALL-CUSHION-cue, BALL-BALL-cue-red, BALL-POCKET-red-rt
3. BALL-CUSHION-cue, BALL-CUSHION-cue, BALL-BALL-cue-red, BALL-POCKET-red-rt
```
or 
```
STRATEGY: none
DIFFICULTY: medium
SHOTS:
1. BALL-BALL-cue-red, BALL-CUSHION-red, BALL-BALL-red-blue, BALL-POCKET-red-rt
2. BALL-BALL-cue-blue, BALL-CUSHION-blue, BALL-POCKET-blue-rb
3. BALL-CUSHION-cue, BALL-BALL-cue-yellow, BALL-POCKET-yellow-lt
```

These are the rules that you MUST follow:
    - Be creative in your choice of events
    - Do not repeat shots
    - DO NOT suggest a shot that fouls by
        1) potting a ball that is not a target ball, 
        2) hitting a non-target ball first, 
        3) or by potting the cue ball.

Use the Reasoning field to briefly think of what makes a good pool shot, and why you are choosing the shot you choose.
\end{lstlisting}

\subsection{Explainer Prompt}
\label{app:lm-prompts:explainer}

The explainer prompts the LM with the task description detailed in Listing~\ref{lst:explainer:td}, and the following inputs:
\begin{itemize}
\item{Shot parameters: $\T=(v, \alpha, \beta, a, b)$ ; }
\item{Balls and Pockets board coordinates of the current state $\X\in\mathcal X$: the exact (x,y) coordinates of each ball and pocket on the table ; }
\item{Events $L\in\mathcal L$: the events that occurred in the shot, and their positions ; }
\item{Value rules description $R^v_i\in\mathcal L$ and evaluations $r^v_i(\X,\T)$: the rules and weights (as percentages) that contribute to the value of the shot, i.e. a high value means the shot is good for the reason stated in the rule, and a low value means the shot is bad for that reason ; }
\item{Difficulty rules descriptions $R^d_i\in\mathcal L$ and evaluations $r^d_i(\X,\T)$: the rules and weights (as percentages) that contribute to the difficulty of the shot, i.e. a high difficulty means the shot is hard for the reason stated in the rule, and a low value means the shot is easy for that reason . }
\end{itemize}

The list of outputs that it needs to return are as follows:
\begin{itemize}
\item{Explanation: the explanation of the shot that takes into account both the value and difficulty of the shot and makes reference to the rules and weights that contribute to the value and difficulty of the shot.}
\end{itemize}

\lstset{caption={Explainer Prompt - Task Description}, label={lst:explainer:td}}
\begin{lstlisting}[escapechar=@]
You are tasked with explaining the pros and cons of a particular shot in a game of pool using the following information:
    - The target balls, i.e. the balls that should be potted, are 'blue', 'red', and 'yellow', and not 'green', 'black', or 'pink'
    - The shot parameters are provided, and are defined as:
        - V0: the initial velocity of the cue ball
        - theta: The angle of inclination of the cue stick
        - phi: The angle of rotation of the cue stick
        - a: The x offset of the cue stick
        - b: The y offset of the cue stick
    - The exact (x,y) coordinates of each ball and pocket on the table
    - The events that occurred in the shot, and their positions
    - The value rules and weights
    - The difficulty rules and weights

You must return an explanation of the pros and cons of the shot, that takes into account both the value and difficulty of the shot, with regard to the rules provided. You are also given how each rule applies to the current state and shot, as a statement:
    - None
    - Low 
    - Medium
    - High
    - Extremely high
Imagine you are explaining to a curious student who wants to learn more about the game of pool. Make it seem natural and conversational, while also thorough and full of detail, and be sure to not refer to numbers of the rules and weights, you must rewrite them into something more natural. Also, be sure to explain any pool specific words used. Above all, keep the explanation short and concise, no more than 10 lines.
\end{lstlisting}

\newpage
\section{Further Details on Relevance to Expert Rules}
\label{app:exp2}

Section $4.2.2$ detailed an experiment to validate the reliability of the explanations provided by \Assistant\, with results presented in Section $5.2.2$.
We present in Listings~\ref{lst:exp2:td:with} and~\ref{lst:exp2:td:without}  the task descriptions employed in conjunction with the template prompt presented in Listing~\ref{lst:template} for the contexts of, respectively, the experimental condition `With $r_r$' and the control condition `Without $r_r$'.

\lstset{caption={Relevance to Expert Rules Prompt - Task Description - With $r_r$}, label={lst:exp2:td:with}}
\begin{lstlisting}[escapechar=@]
You are tasked with providing estimations of the applicability of some value and difficulty rules to a particular shot in a game of pool, using the following information:
    - The target balls, i.e. the balls that should be potted, are 'blue', 'red', and 'yellow'
    - The shot parameters are provided, and are defined as:
        - V0: the initial velocity of the cue ball
        - theta: The angle of inclination of the cue stick
        - phi: The angle of rotation of the cue stick
        - a: The x offset of the cue stick
        - b: The y offset of the cue stick
    - The exact (x,y) coordinates of each ball and pocket on the table
    - The events that occurred in the shot, and their positions
    - The value rules and weights
    - The difficulty rules and weights

You must return an estimate of the applicability of each of the value and difficulty rules.
The value and difficulty rules' weights are percentages that represent the extent to which a given rule applies to the current state and shot.
To convert a given rule's percentage weight X into a valid estimation of the applicability of the given rule to the current state and shot, proceed as follows:
    - if 0 <= X < 12.5, then the given rule's applicability is very low ;
    - if 12.5 <= X < 25, then the given rule's applicability is low ;
    - if 25 <= X < 37.5, then the given rule's applicability is moderately low ;
    - if 37.5 <= X < 62.5, then the given rule's applicability is moderate ;
    - if 62.5 <= X < 75, then the given rule's applicability is moderately high ;
    - if 75 <= X < 87.5, then the given rule's applicability is high ;
    - if 87.5 <= X <= 100, then the given rule's applicability is very high.
\end{lstlisting}

\lstset{caption={Relevance to Expert Rules Prompt - Task Description - Without $r_r$}, label={lst:exp2:td:without}}
\begin{lstlisting}[escapechar=@]
You are tasked with providing estimations of the applicability of some value and difficulty rules to a particular shot in a game of pool, using the following information:
    - The target balls, i.e. the balls that should be potted, are 'blue', 'red', and 'yellow'
    - The shot parameters are provided, and are defined as:
        - V0: the initial velocity of the cue ball
        - theta: The angle of inclination of the cue stick
        - phi: The angle of rotation of the cue stick
        - a: The x offset of the cue stick
        - b: The y offset of the cue stick
    - The exact (x,y) coordinates of each ball and pocket on the table
    - The events that occurred in the shot, and their positions
    - The value rules
    - The difficulty rules

You must return an estimate of the applicability of each of the value and difficulty rules.
\end{lstlisting}

The following inputs are provided to the LM:
\begin{itemize}
\item{Shot parameters: $\T=(v, \alpha, \beta, a, b)$ ;}
\item{Balls \& Pockets board coordinates of the current state $\X\in\mathcal X$: the exact (x,y) coordinates of each ball and pocket on the table ;}
\item{
Events $L\in\mathcal L$: the events that occurred in the shot, and their positions ;}
\item{Value rules description $R^v_i\in\mathcal L$ and, only in the case of the experimental condition `With $r_r$', the evaluations $r^v_i(\X,\T)$: the rules and weights (as percentages) that contribute to the value of the shot, i.e. a high value means the shot is good for the reason stated in the rule, and a low value means the shot is bad for that reason ;}
\item{Difficulty rules descriptions $R^d_i\in\mathcal L$ and, only in the case of the experimental condition `With $r_r$', evaluations $r^d_i(\X,\T)$: the rules and weights (as percentages) that contribute to the difficulty of the shot, i.e. a high difficulty means the shot is hard for the reason stated in the rule, and a low value means the shot is easy for that reason .}
\end{itemize}

The list of outputs that the LM is prompted to return consists of an $m$-vector of strings containing the corresponding Likert values on the Likert scale detaild in Section $4.2.2$.

\begin{figure*}[ht]
\centering
\includegraphics[width=\textwidth]{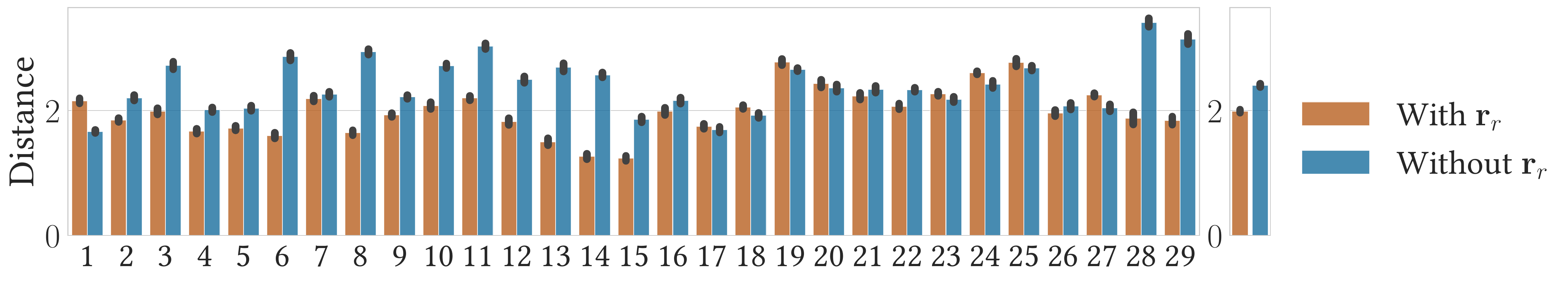}
\caption{\textbf{Left:} Distributions of Likert scale distances between ground truth and estimations from different methods for each domain expert rule (mean$\pm$std. error) ; \textbf{Right:} Aggregated results over all domain expert rules (mean$\pm$std. error) ; Results obtained using the language model \textit{Llama-3.2-3B-Instruct}~\citep{llama3modelcard,llama-3-3B}.}
\label{fig:exp2-3B}
\end{figure*}

\begin{figure*}[ht]
    \centering
    \includegraphics[width=\textwidth]{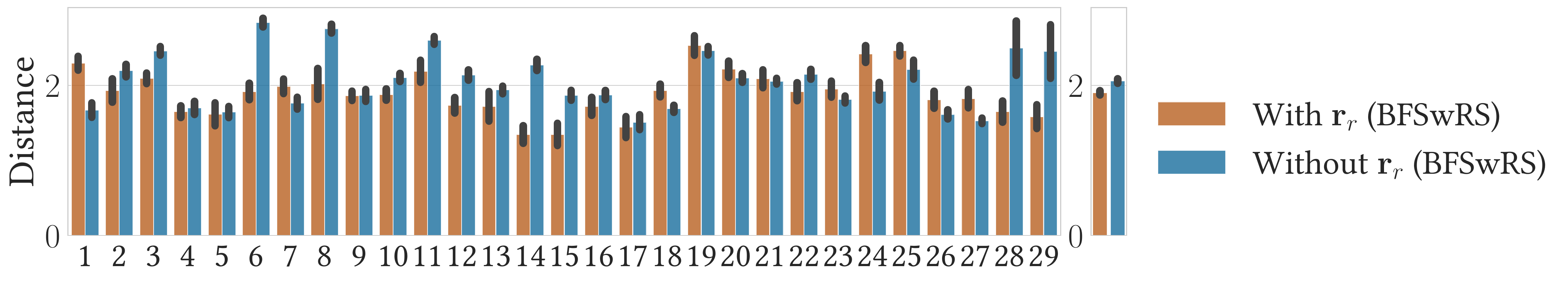}
    \caption{
    \textbf{Left:} Distributions of Likert scale distances between ground truth and estimations from different methods for each domain expert rule (mean$\pm$std. error) 
    ;
    \textbf{Right:} Aggregated results over all domain expert rules (mean$\pm$std. error) ;
	Results obtained using the language model \textit{Llama-3.2-3B-Instruct}~\citep{llama3modelcard,llama-3-3B}, after prompt optimisation using DSPy~\citep{khattab2024dspy}'s `BootstrapFewShotwithRandomSearch` (BFSwRS) with maximum number of few-shot examples $n=4$.
    }
    \label{fig:exp2-3B-BFSwRS}
\end{figure*}

\begin{figure*}[ht]
    \centering
    \includegraphics[width=\textwidth]{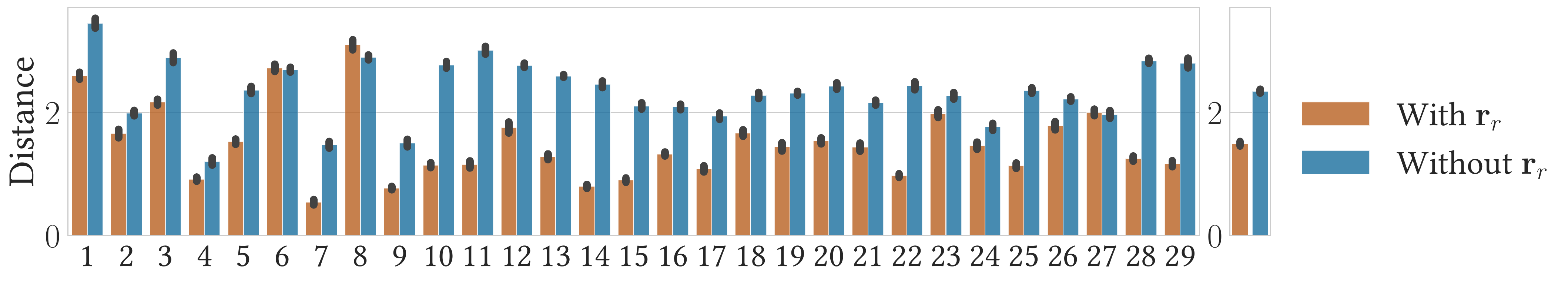}
    \caption{
    \textbf{Left:} Distributions of Likert scale distances between ground truth and estimations from different methods for each domain expert rule (mean$\pm$std. error) 
    ;
    \textbf{Right:} Aggregated results over all domain expert rules (mean$\pm$std. error) ;
    Results obtained using the language model \textit{Llama-3.1-8B-Instruct}~\citep{llama3modelcard,llama-3-8B}.
    }
    \label{fig:exp2-8B}
\end{figure*}

\begin{figure*}[ht]
    \centering
    \includegraphics[width=\textwidth]{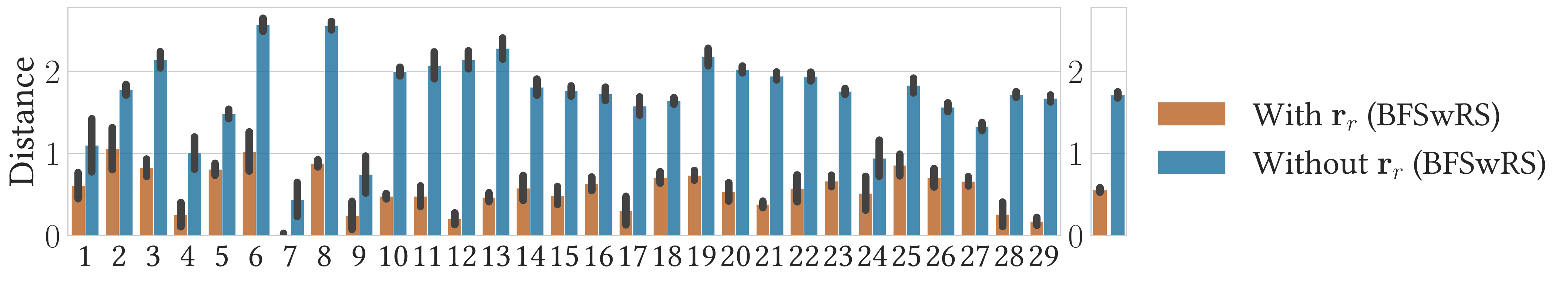}
    \caption{
    \textbf{Left:} Distributions of Likert scale distances between ground truth and estimations from different methods for each domain expert rule (mean$\pm$std. error) 
    ;
    \textbf{Right:} Aggregated results over all domain expert rules (mean$\pm$std. error) ;
	Results obtained using the language model \textit{Llama-3.1-8B-Instruct}~\citep{llama3modelcard,llama-3-8B}, after prompt optimisation using DSPy~\citep{khattab2024dspy}'s `BootstrapFewShotwithRandomSearch` (BFSwRS) with maximum number of few-shot examples $n=4$.
    }
    \label{fig:exp2-8B-BFSwRS}
\end{figure*}

\begin{figure*}[ht]
    \centering
    \includegraphics[width=\textwidth]{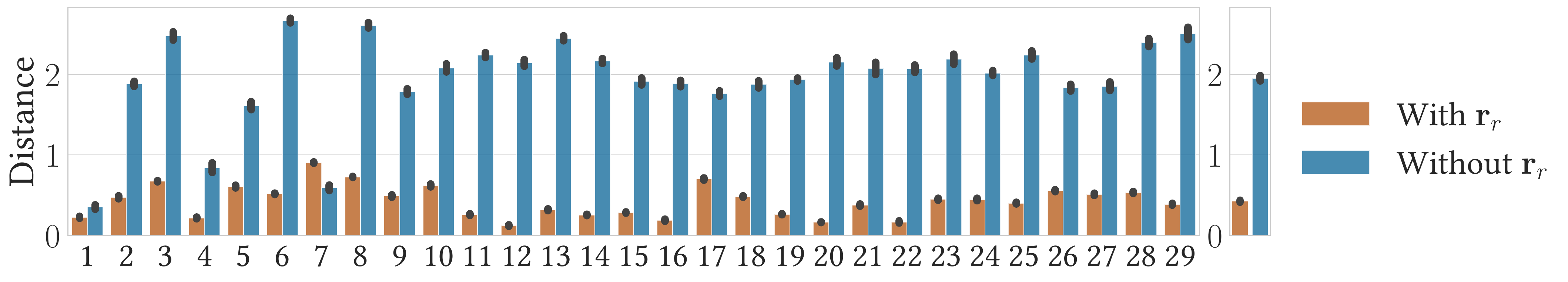}
    \caption{
    \textbf{Left:} Distributions of Likert scale distances between ground truth and estimations from different methods for each domain expert rule (mean$\pm$std. error) 
    ;
    \textbf{Right:} Aggregated results over all domain expert rules (mean$\pm$std. error) ;
    Results obtained using the language model \textit{Llama-3.1-70B-Instruct}~\citep{llama3modelcard,llama-3-70B}.
    }
    \label{fig:exp2-70B}
\end{figure*}

\begin{figure*}[ht]
    \centering
    \includegraphics[width=\textwidth]{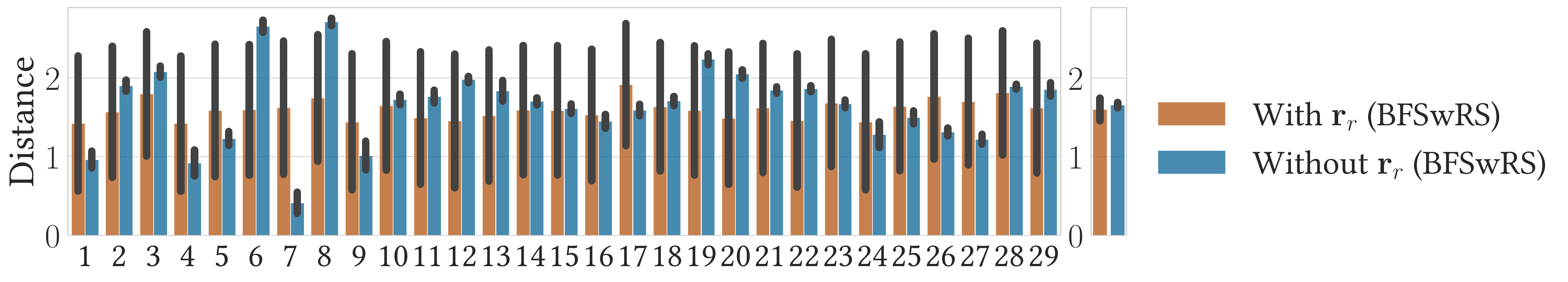}
    \caption{
    \textbf{Left:} Distributions of Likert scale distances between ground truth and estimations from different methods for each domain expert rule (mean$\pm$std. error) 
    ;
    \textbf{Right:} Aggregated results over all domain expert rules (mean$\pm$std. error) ;
	Results obtained using the language model \textit{Llama-3.1-70B-Instruct}~\citep{llama3modelcard,llama-3-70B}, after prompt optimisation using DSPy~\citep{khattab2024dspy}'s `BootstrapFewShotwithRandomSearch` (BFSwRS) with maximum number of few-shot examples $n=4$.
    }
    \label{fig:exp2-70B-BFSwRS}
\end{figure*}

The discussion in Section $6$ highlights that performance on the task scales positively with model size, as evidenced by decreasing Likert distance bars between model estimations and ground truth values in Figures~\ref{fig:exp2-3B},~\ref{fig:exp2-8B} and ~\ref{fig:exp2-70B}, respectively for model sizes small (3B), medium (8B), and large (70B).
In order to guard against the confounder of prompt quality, we perform additional experiments with prompt optimization using DSPy~\cite{khattab2024dspy}'s `BootstrapFewShotwithRandomSearch` (BFSwRS) with maximum number of few-shot examples $n=4$. 
Results show statistically significant improvements only for the medium-sized model, as seen in Figures~\ref{fig:exp2-3B-BFSwRS},~\ref{fig:exp2-8B-BFSwRS}, and~\ref{fig:exp2-70B-BFSwRS}, ordered in increasing model size. 

\newpage
\pagebreak
\section{Details about Teaser Figure}
\label{app:teaser}

We present in Listing~\ref{lst:teaser:details} the user query that lead to the shot and explanations presented in Figure 1 of the main paper, along with clarification of the positive and negative rules.

\lstset{caption={Details about Teaser Figure}, label={lst:teaser:details}}
\begin{lstlisting}[escapechar=@]
**USER QUERY**
 
Pot a ball and leave the cue ball in a defensive position
  
**EXPLANATION**
   
This shot is good if you are feeling confident about your ball control and want to play defensively, though you will need to get your speed and spin just right. Fortunately, even if you miss potting the ball you are leaving the cue ball in a safe position, putting your opponent in a tough spot for their next shot.

**POSITIVE RULES**
     
5: Safety Opportunities: Identify chances to play defensive shots that leave the opponent in a difficult position. Good safety opportunities can be as valuable as offensive shots in many situations.
      
6: Two-way Shot Possibilities: Look for shots that offer both offensive and defensive potential. These shots allow for pocketing a ball while also setting up a good defensive position if missed, providing strategic flexibility.

13: Above all, prioritise shots that pot the most target balls.

**NEGATIVE RULES**
	 
18: English Requirements: Shots needing sidespin (English) are more difficult to control and execute. The use of English introduces additional variables that affect both aim and cue ball behavior after contact.

19: Speed Control: Shots requiring precise speed control, whether very fast or very slow, are more challenging. Speed affects pocket geometry, throw, and positioning for the next shot.
\end{lstlisting}

\newpage
\pagebreak
\section{User Study}
\label{app:user-study}

\begin{figure*}[ht]
\includegraphics[width=\textwidth]{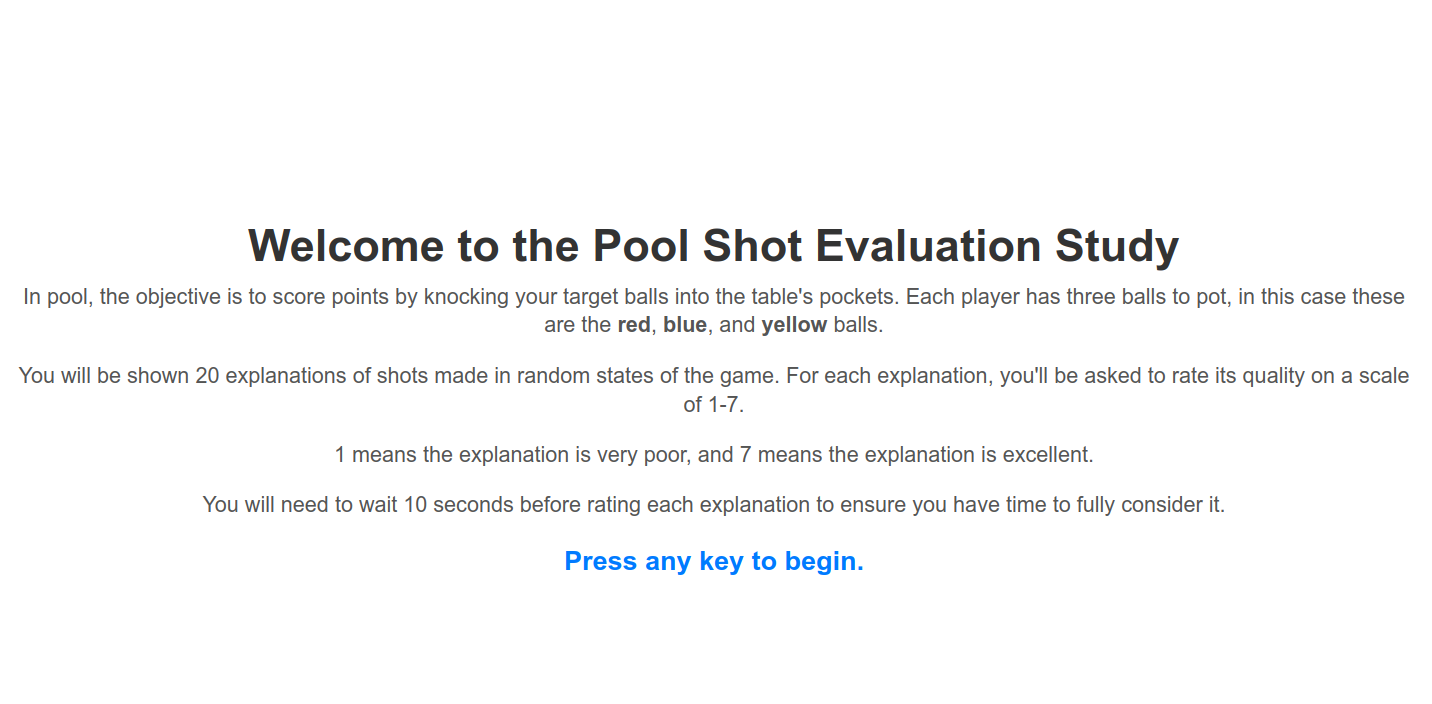}
\caption{
Landing page of the user study from Section $4.2.3$.
}
\label{fig:user-study:landing}
\end{figure*}

\begin{figure*}[h]
\includegraphics[width=0.8\textwidth]{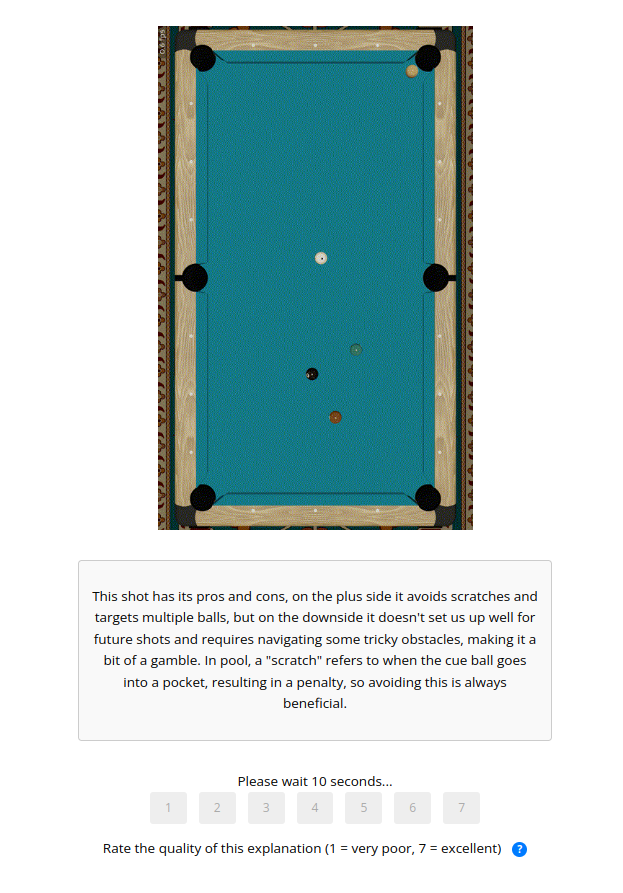}
\caption{
Example of prompt to user of the user study from Section $4.2.3$.
}
\label{fig:user-study:example}
\end{figure*}

Figures~\ref{fig:user-study:landing} and~\ref{fig:user-study:example} shows, respectively, the landing page of the user study of Section $4.2.3$, and an example of a prompt to user with an explanation of a shot and its related GIF visualisation.

